\documentclass[nonacm,screen,oneside]{acmart}

\usepackage[utf8]{inputenc}
\usepackage[T1]{fontenc}
\usepackage[capitalise,noabbrev,nameinlink]{cleveref}
\crefformat{equation}{#2Equation~#1#3}
\crefmultiformat{equation}{Equations~#2#1#3}{ and~#2#1#3}{, #2#1#3}{ and~#2#1#3}

\usepackage{tikz}
\usetikzlibrary{backgrounds,calc,arrows,arrows.meta,positioning}
\makeatletter\tikzset{add font/.code={\expandafter\def\expandafter\tikz@textfont\expandafter{\tikz@textfont#1}}}\makeatother
\tikzset{every node/.style={add font=\sffamily}}
\tikzset{every picture/.style={tight background,>=stealth}}

\usepackage{enumitem}
\usepackage[english]{babel}
\usepackage[normalem]{ulem}

\acmConference[Conference'21]{ACM Conference}{}{}

\DeclareMathOperator*{\argmin}{argmin}
\newcommand{\abs}[1]{\left\lvert#1\right\rvert}
\newcommand{\norm}[1]{\left\lVert#1\right\rVert}

\begin{document}
	
\title{Real-Time Global Illumination Decomposition of Videos}

\author{Abhimitra Meka}
\affiliation{%
	\institution{Max Planck Institute for Informatics, Saarland Informatics Campus and Google}\country{Germany}}
\email{abhim@google.com}
\authornote{Authors contributed equally to this work.}
\author{Mohammad Shafiei}
\affiliation{%
	\institution{Max Planck Institute for Informatics, Saarland Informatics Campus}\country{Germany}}
\email{moshafie@eng.ucsd.edu}
\authornotemark[1]
\author{Michael Zollh\"{o}fer}
\affiliation{%
	\institution{Stanford University}\country{USA}}
\email{zollhoefer@stanford.edu}
\author{Christian Richardt}
\affiliation{%
	\institution{University of Bath}\country{UK}}
\email{christian@richardt.name}
\author{Christian Theobalt}
\affiliation{%
	\institution{Max Planck Institute for Informatics, Saarland Informatics Campus}
	\streetaddress{Campus E1.4}
	\city{Saarbrücken}
	\postcode{66123}
	\country{Germany}}
\email{theobalt@mpi-inf.mpg.de}

\renewcommand\shortauthors{Meka, A., Shafiei, M., et al.}

\begin{abstract}
We propose the first approach for the decomposition of a monocular color video into direct and indirect illumination components in real time.
We retrieve, in separate layers, the contribution made to the scene appearance by the scene reflectance, the light sources and the reflections from various coherent scene regions to one another.
Existing techniques that invert global light transport require image capture under multiplexed controlled lighting, or only enable the decomposition of a single image at slow off-line frame rates.
In contrast, our approach works for regular videos and produces temporally coherent decomposition layers at real-time frame rates.
At the core of our approach are several sparsity priors that enable the estimation of the per-pixel direct and indirect illumination layers based on a small set of jointly estimated base reflectance colors.
The resulting variational decomposition problem uses a new formulation based on sparse and dense sets of non-linear equations that we solve efficiently using a novel alternating data-parallel optimization strategy.
We evaluate our approach qualitatively and quantitatively, and show improvements over the state of the art in this field, in both quality and runtime.
In addition, we demonstrate various real-time appearance editing applications for videos with consistent illumination.

\end{abstract}

\begin{teaserfigure}
	\includegraphics[width=\textwidth]{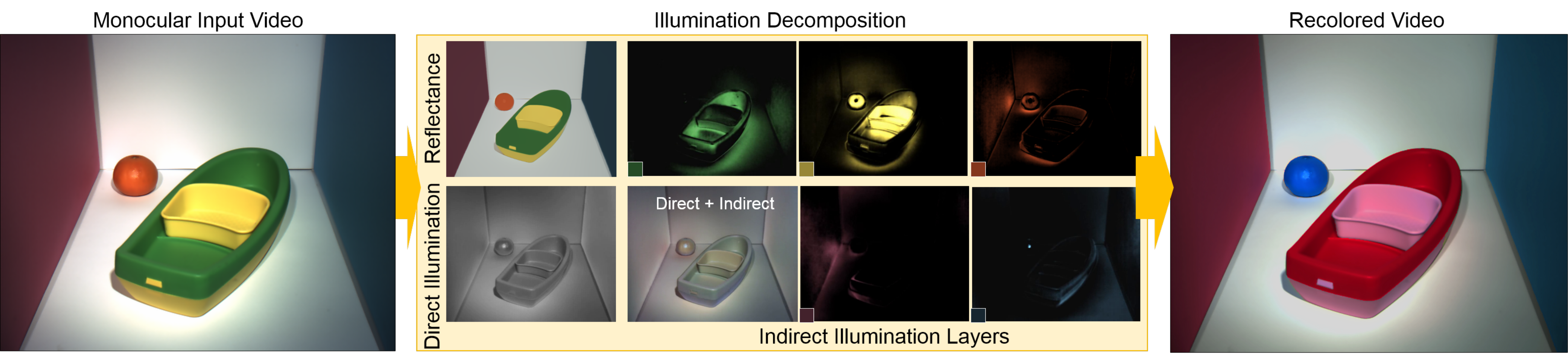}
	\caption
	{
		We propose the first approach for the real-time decomposition of a video into direct and indirect illumination components.
		Our approach decomposes a video (left) into its reflectance, direct illumination, and multiple indirect illumination layers (middle) that explain the light transport in the scene up to the first bounce.
		This enables various real-time appearance editing applications with interactive user feedback, such as inter-reflection consistent recoloring (right).
	}
	\label{fig:teaser}
\end{teaserfigure}


\maketitle


\section{Introduction}
\label{sec:introduction}

The appearance of each pixel in a real-world image is the combined result of complex light and material interactions that can be mathematically described by the rendering equation~\cite{Kajiy1986}.
While the rendering equation models the radiance (the light energy radiated outwards) of a surface point, it is also a function of the irradiance (the light energy incident) on the surface point.
In a scene with complex geometry, one point's radiance could be a distant point's irradiance.
This leads to a complex set of back-and-forth interactions of light reflections, known as global illumination, that define the appearance of the pixels.

Understanding these global illumination effects is crucial to appearance editing applications, as modifying the appearance of one region of the frame has an effect on other regions (see example in \cref{fig:teaser}).
Solving for these interactions is an underconstrained problem of decomposing each pixel into the components of light transport, light distribution, or materials in a scene, all without knowing the geometry. %
This creates intriguing new possibilities in increasingly important image and video editing applications, and in augmented reality.
This also has the potential to stabilize more general computer vision algorithms under difficult illumination.
Classically, techniques that attempt to invert the phenomenon of light transport in a scene and retrieve the various transmission and reflection components have relied on multi-step active illumination projector and camera systems \cite{SeitzMK2005, NayarKGR2006}.
Although such systems accurately separate the direct illumination  from the global lighting components, they still do not efficiently characterize the appearance inter-dependence between the various points in the scene.
Hence, such a decomposition does not enable editing applications which require manipulation of specific scene regions.

Recently, \citet{DongDTP2015} developed a unique representation of light transport that allows for acquiring a low number of projection-acquisition image pairs which can be efficiently utilized to derive various intrinsic reflection components between scene regions, thus better encoding the interdependence of surface appearance.
Using this technique, they were able to demonstrate globally consistent appearance editing applications.
Yet, their method is encumbered by the hardware and acquisition requirements, making it impossible to be applied to existing images or videos.

In contrast, recent image-based methods solve a color unmixing problem with a sparse set of base colors to decompose an RGB image into layers that can be manipulated independently.
\citet{AksoyAPS2016} solve the color unmixing along with a matting problem without computing interpretable layers such as scene reflectance or shading.
\citet{CarroRA2011} first compute a two-layer intrinsic image decomposition using the user-interactive method of \citet{BoussPD2009}, and then solve the color unmixing problem on the shading image alone.

While these methods are significant steps towards decomposing light transport in images, the problem of decomposing live videos, which is more widely applicable, still remains a challenge. 
Inspired by the sparse base color assumption, we present the first method to perform a fully temporally coherent decomposition of a video into scene reflectance, a direct illumination layer and multiple indirect illumination layers, at real-time frame rates.
The direct illumination layer represents the contribution made directly by the light source to the scene radiance, and the indirect illumination layers encode the contribution that one region of the scene makes to the radiance of other regions.
We show that the indirect illumination has a natural sparsity which is a useful tool in estimating the illumination layers, and also in refining the scene reflectance.
In summary, the core algorithmic novelties, in addition to the real-time system processing live videos, that distinguish our work from previous approaches are:
\begin{enumerate}
\item Joint illumination decomposition of direct and indirect illumination layers, and estimation and refinement of base colors that constitute the scene reflectance.

\item A sparsity-based automatic estimation of the underlying reflectance when a user identifies regions of strong inter-reflections.

\item A novel parallelized sparse–dense optimizer to solve a mixture of high-dimensional sparse problems jointly with low-dimensional dense problems at real-time frame rates.
\end{enumerate}
Based on our decomposition, we show appearance editing applications on videos, and demonstrate qualitative and quantitative improvements over the state of the art.


\section{Related Work}
\label{sec:relatedwork}

\paragraph{Inverse Rendering}

The colors in an image depend on scene geometry, material appearance and illumination.
Reconstructing these components from a single image or video is a challenging and ill-posed problem called \emph{inverse rendering} \citep{YuDMH1999,RamamH2001,PatowP2003}.
Most approaches need to make strong assumptions to estimate material and illumination, such as the
availability of an RGBD camera \citep[e.g.][]{WuWZ2016,GuoXYLDL2017}, strong priors such as a data-driven BRDF model \citep{LombaN2016} or flash lighting \citep{LiXRSC2018,NamLGK2018},
knowledge of geometry \citep{MarscG1997,DongCPZT2014,LiDPT2017,AzinoLKN2019}
or a specific object class \citep{GeorgRRGFVT2018,LiuCYYL2017}.
As we will show, many complex image editing tasks can be achieved using a purely image-based decomposition without full inverse rendering of the above-mentioned kind.

\paragraph{Global Illumination Decomposition}

To decompose the captured radiance of a scene into direct and indirect components, some methods actively illuminate the scene to investigate the effect of light transport.
\citet{SeitzMK2005} use a laser to sequentially light up the corresponding geometry of each pixel, and \citet{NayarKGR2006} and \citet{OToolMK2016} use multiple images captured under structured lighting.
While these methods use active illumination to decompose scene radiance into direct and indirect components, they cannot separate reflectance and illumination.
Thus, these methods cannot ascertain which object causes which color spill, which makes applications such as recoloring or material editing impossible.
On the other hand, \citet{DongDTP2015} estimate the global illumination caused by diffuse regions of interest, which allows them to perform recoloring on those regions with consistent light interactions with the scene.
\citet{LaffoBPDD2012} proposed an approach for intrinsic decomposition based on a photo collection of a scene under different viewpoints/illuminations to better constrain the problem.
\citet{RenDLTG2015} propose a data-driven method for image-based rendering of a scene under novel illumination conditions by taking multiple images of the same scene with different illumination settings as input.
\citet{YuDMH1999} estimate the diffuse and specular reflectance map as well as indirect illumination.
To this end, they solve inverse radiosity by taking multiple calibrated HDR images with known direct illumination as input along with the geometry of the scene.
Our approach only requires a single color image or video to estimate the direct reflectance and illumination – in addition to decomposing the indirect illumination.

\begin{figure*}
	\includegraphics[width=\linewidth]{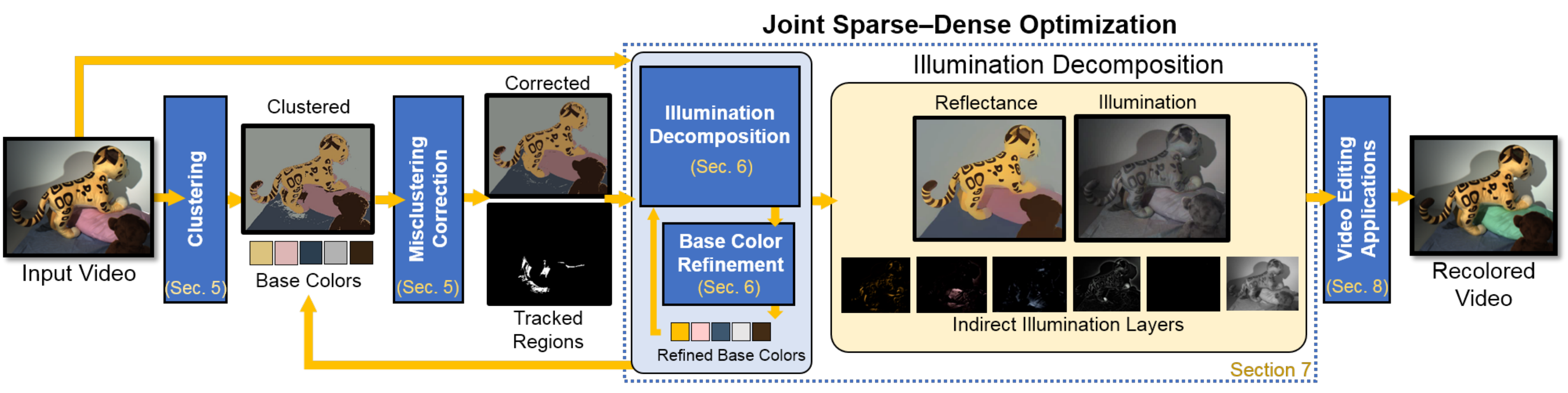}
	\caption{\label{fig:pipeline}%
		Given a monocular color video as input, our approach estimates the light transport decomposition at real-time frame rates.
		At the core of our approach are several sparsity priors that enable the estimation of per-pixel direct and indirect decomposition layers based on a small set of jointly estimated base colors.
		The resulting variational problem is efficiently solved using a novel alternating data-parallel optimization strategy.
		The decomposition is the basis for several compelling live video editing applications, such as inter-reflection consistent recoloring.
	}
\end{figure*}

\paragraph{Intrinsic Images}

Many approaches have been introduced for the task of intrinsic image decomposition that explains a photograph using physically interpretable images such as reflectance and shading \citep{BarroT1978}; see \citet{BonneKPB2017} for a recent survey.
Given the challenging ambiguity of such a decomposition, most methods impose the assumption of white illumination by constraining the shading image to be grayscale
\citep{ZoranIKF2015,BiHY2015,ZhouKE2015,BellBS2014,BonneSTSPP2014,YeGLDG2014,MekaZRT2016,DingSHXM2017,JanneWKYT2017,KovacBSB2017},
while few methods support a colored shading layer
\citep{BarroM2015,ChangCF2014,BoussPD2009,KimPSL2016,ShiDSY2017}.
%
Colored shading effects can result either from a colored light source or global illumination effects such as inter-reflections.
Due to the ill-posedness of the intrinsic decomposition task – particularly with non-white illumination – some methods require object segmentation \citep{BeigpW2011} or significant user guidance \citep{BoussPD2009, ShenYJL2011} proportional to the complexity of the input image.
Although we assume a white illuminant, we represent our illumination layer using RGB to capture the colored inter-reflections between objects.
We take inspiration from the locally constrained clustering approach of \citet{GarceMLG2012}, which segments the image in Lab color space based on chroma variations using k-means clustering, but has slow off-line run times.
Like \citet{MekaZRT2016}, we perform clustering using a histogram to reduce the run time. Methods working with light-field images enforce reflectance and shading consistency across multiple views to perform better intrinsic image decomposition \citep{AlperG2017,GarceEZWZG2017}.
However, this is not applicable to monocular videos.
Several recent methods use learning-based approaches to perform intrinsic image decomposition.
The methods of \citet{NarihMY2015b} and \citet{ShiDSY2017} use synthetic data to train their neural networks to perform an intrinsic decomposition task, whereas \citet{NestmG2017} use ordinal relationship of reflectance values in the image to train their network, based on the Intrinsic Images in the Wild dataset \citep{BellBS2014}.
\citet{BiKR2018} use a hybrid approach that utilizes both synthetic and real training data.
Our method is significantly different from these data-driven approaches because we do not rely on any large datasets to learn our priors.

\paragraph{Intrinsic Video}

The intrinsic decomposition task is even more challenging for videos.
Naïvely decomposing every video frame leads to a temporally incoherent decomposition.
Therefore, \citet{KongGB2014}, \citet{BonneTSSPP2015} and \citet{MekaZRT2016} employ temporal consistency priors, and \citet{BonneSTSPP2014} and \citet{YeGLDG2014} use an optical-flow based consistency constraint.
\citet{ShenYCSL2014} estimate the intrinsic decomposition only for a specific region and thus require user input.
Among these methods, only \citet{BonneSTSPP2014} and \citet{MekaZRT2016} can perform more than one decomposition per second, with the latter achieving real-time frame rates.
In our approach, we solve a more challenging problem that requires a higher number of parameters: direct reflectance and illumination, and multiple indirect illumination layers – all in real time.
The underlying optimization problem exhibits a mixed sparse–dense structure, which makes current data-parallel GPU solvers \citep{ZollhNIRZFWFLTS2014,MekaZRT2016,DeVitMZBRTHFN2017} inefficient.
We tackle this problem using a sparse–dense splitting strategy that leads to higher throughput.
We also integrate the possibility for user strokes into our system to better disambiguate between the reflectance and illumination layers.
These annotations are automatically propagated across all video frames (see \cref{sec:misclustering}).

\paragraph{Layer-based Image Editing}

A physically accurate decomposition is not required to achieve complex image editing tasks such as recoloring of objects.
Instead, a decomposition into multiple semitransparent layers is often sufficient, as demonstrated for instance by image vectorization techniques \citep{RichaLBAD2014,FavreLB2017}.
\citet{AksoyAPS2016} introduce an interactive color unmixing approach that decomposes an image or video into additive layers of dominant scene colors.
This enables accurate green-screen keying and layer recoloring, but requires a user to manually identify all base colors.
\citet{TanLG2016} automatically estimate a given number of base colors using the vertices of the simplified convex hull of observed RGB colors.
However, the user still needs to determine the order of the layers.
\citet{AksoyASP2017} determine the base color model fully automatically, and then decompose images into high-quality, additive, near-uniformly colored layers.
They demonstrate a large variety of layer adjustments that are enabled by their decomposition.
\citet{InnamRWM2017} learn an image decomposition into a mixture of additive and multiplicative layers for occlusion, albedo, irradiance and specular layers, instead of layers of distinct colors.
\citet{LinFDH2017} represent each pixel in an image by a linear combination of base colors and nearest neighbors.
The former combination enables color editing and the latter allows soft color blending.
\citet{TanEG2018a} perform additive decomposition in real time given a fixed palette of base colors.
We combine intrinsic decomposition with layer-based decomposition of the illumination layer that enables new video editing tasks that go beyond those supported by existing layer-based decompositions of images.

\section{Overview}
\label{sec:overview}

We present the first real-time method for temporally coherent illumination decomposition of a video into a reflectance layer, direct illumination layer and multiple indirect illumination layers.
\Cref{fig:pipeline} shows the major components of our method and how they interact.
We propose a novel sparsity-driven formulation for the estimation and refinement of a base color palette, which is used for decomposing the video frames (see \cref{sec:formulation}).
Our algorithm starts by automatically estimating a set of base colors that represent scene reflectances (see \cref{sec:basecolorestimation}).
Unlike previous methods that heavily rely on user interaction, our method is automatic and only occasionally requires a minimal set of user clicks on the first video frame to identify regions of strong inter-reflections and refine the base colors.
We propagate the user input automatically to the rest of the video by a spatiotemporal region-growing method.
We then perform the illumination decomposition (see \cref{sec:energy}).
Our formulation results in a mixture of dense and sparse non-convex high-dimensional optimization problems, which we solve efficiently using a custom-tailored parallel iterative non-linear solver that we implement on the GPU (see \cref{sec:optimization}).
We show that our optimization technique achieves real-time frame rates on modern commodity graphics cards.

We evaluate our method on a variety of synthetic and real-world scenes, and provide comparisons that show that our method outperforms state-of-the-art illumination decomposition, intrinsic decomposition and layer-based image editing techniques, both qualitatively and quantitatively (see \cref{sec:results}).
We also demonstrate that real-time illumination decomposition of videos enables a range of advanced, illumination-aware video editing applications that are suitable for photo-real augmented reality applications, such as inter-reflection-aware recoloring and retexturing (see \cref{sec:applications}).


\section{Problem Formulation}
\label{sec:formulation}

Our algorithm decomposes each video frame into a reflectance layer, a direct illumination layer and multiple indirect illumination layers.
In order to achieve such a decomposition, we make some simplifying assumptions about the scene, as listed below:
\begin{itemize}
  \item We assume that the scene is Lambertian, i.e., surfaces exhibit no view-dependent effects and hence their reflectance can be parameterized as a diffuse albedo with RGB components.
  \item We assume that all light sources in the scene produce only white colored light. Hence, the direct illumination in the scene can be expressed by a grayscale or single channel image.
  \item We assume that the second reflection bounce (or the first inter-reflection) of light is the primary source of indirect illumination in the scene, while the contribution of subsequent bounces of light is negligible.
  \item We assume that the motion of the camera in the video is smooth with significant overlap between adjacent frames.
  \item We also assume that no new objects or materials come into view after the first frame. 
\end{itemize}
These assumptions allow us to have a linear formulation for the light transport in the scene, as discussed in this section. Please note that the first three assumptions are also made by the current state-of-the-art approaches of \citet{CarroRA2011} and \citet{MekaZRT2016}. Only the last two assumptions are specific to our method and are reasonable assumptions about the nature of the captured video.

Our algorithm factors each video frame $\mathbf{I}$ into a per-pixel product of the reflectance $\mathbf{R}$ and the illumination $\mathbf{S}$:
\begin{equation}
\label{eq:intrinsic}
\mathbf{I}(\mathbf{x}) = \mathbf{R}(\mathbf{x}) \odot \mathbf{S}(\mathbf{x}) \text{,}
\end{equation}
where $\mathbf{x}$ denotes the pixel location and $\odot$ the element-wise product.
For diffuse objects, the reflectance layer captures the surface albedo, and the illumination layer $\mathbf{S}$ jointly captures the direct and indirect illumination effects. 
Unlike most intrinsic decomposition methods, we do not use a grayscale illumination image, but represent the illumination layer as a colored RGB image to allow indirect illumination effects to be expressed in the illumination layer.

Inspired by \citet{CarroRA2011}, we further decompose the illumination layer into a grayscale direct illumination layer resulting from the white illuminant, and multiple indirect colored illumination layers resulting from inter-reflections from colored objects in the scene.
We start by estimating a set of base colors that consists of $K$ unique reflectance colors $\{\mathbf{b}_k\}$ that represent the scene.
The number $K$ of colors is specified by the user; we use $K \!=\! 10$ for all our results, as superfluous clusters will be removed automatically in \cref{sec:clustering}.
This set of base colors serves as the basis for our illumination decomposition.
The base colors help constrain the values of pixels in the reflectance layer $\mathbf{R}$.
For every surface point in the scene, we assume that a single indirect bounce of light may occur from every base reflectance color, in addition to the direct illumination.
The global illumination in the scene is modeled using a linear decomposition of the illumination layer $\mathbf{S}$ into a direct illumination layer $T_0$ and the sum of the $K$  indirect illumination layers $\{T_k\}_{0 < k \leq K}$:
\begin{equation}
\label{eq:imageformation}
\mathbf{I}(\mathbf{x}) = \mathbf{R}(\mathbf{x}) \odot \sum_{k=0}^{K}{ \mathbf{b}_k T_k(\mathbf{x})} \text{.}
\end{equation}
Here, $\mathbf{b}_0$ represents the color of the illuminant: white in our case, i.e. $\mathbf{b}_0 \!=\! (1,1,1)$.
$T_0(\mathbf{x})$ indicates the light transport contribution from the direct illumination.
Similarly, the contribution from each base color $\mathbf{b}_k$ at a given pixel location $\mathbf{x}$ is measured by the map $T_k(\mathbf{x})$.
This scalar contribution, when multiplied with the base color $\mathbf{b}_k$, provides the net contribution by the base reflectance color to the global scene illumination.
Unlike previous methods, we obtain the set of base colors automatically using a real-time clustering technique.
Once the base colors are obtained, the scene clustering can be further refined using a few simple user-clicks.
This refines only the regions of clustering but not the base colors themselves.

In the following sections, we describe the algorithmic steps to estimate and refine the set of base colors and decompose the input video into the set of global illumination layers.

\section{Base Color Estimation}
\label{sec:basecolorestimation}

We initialize the set of base colors by clustering the dominant colors in the first video frame (\cref{sec:clustering}).
This clustering step not only provides an initial base color estimate, but also a segmentation of the video into regions of approximately uniform reflectance.
If needed, the clustering in a video frame undergoes a user-guided correction (\cref{sec:misclustering}).
The base colors are used for the illumination decomposition (\cref{sec:GID}), where they are further refined (\cref{sec:basecolorrefinement}) and used to compute the direct and indirect illumination layers. 

\subsection{Chromaticity Clustering}
\label{sec:clustering}

We cluster the first video frame by color to approximate the regions of uniform reflectance that are observed in scenes with sparsely colored objects.
The locally constrained clustering approach of \citet{GarceMLG2012} segments the image in Lab color space based on chroma variations using k-means clustering, but has slow, off-line run times.
In contrast, our approach is based on a much faster histogram-based k-means clustering approach \citep{MekaZRT2016}.
We perform the clustering of each RGB video frame in a discretized chromaticity space, which makes the clustering more efficient to compute.

The chromaticity image $\mathbf{C}(\mathbf{x}) \!=\! \mathbf{I}(\mathbf{x}) / \abs{\mathbf{I}(\mathbf{x})}$ is obtained by dividing the input image by its intensity \citep{MekaZRT2016,BonneSTSPP2014}.
We then compute a histogram of the chromaticity image with 10 partitions along each axis.
Next, we perform weighted k-means clustering to obtain cluster center chromaticity values, using the population of the bins as the weight and the mid-point of the bin as sample values. 
The user provides an upper limit of the number of clusters visible in the scene ($K \!=\! 10$).
We collapse adjacent similar clusters by measuring the pairwise chromaticity distance between estimated cluster centers.
If this distance is below a threshold of 0.2, we merge the smaller cluster into the larger cluster.
The average RGB colors of all pixels assigned to each cluster then yield the set of initial base colors.
Such a histogram-based clustering approach significantly reduces the segmentation complexity, independent of the image size.
The clustering also produces a segmentation of the input frame, by assigning each pixel to its closest cluster.
This provides a coarse approximation of the reflectance layer, $\mathbf{R}_\mathrm{cluster}$, which we use as an initialization for the reflectance layer $\mathbf{R}$ in the energy optimization detailed in \cref{sec:energy}.

\begin{figure}
	\includegraphics[width=\linewidth]{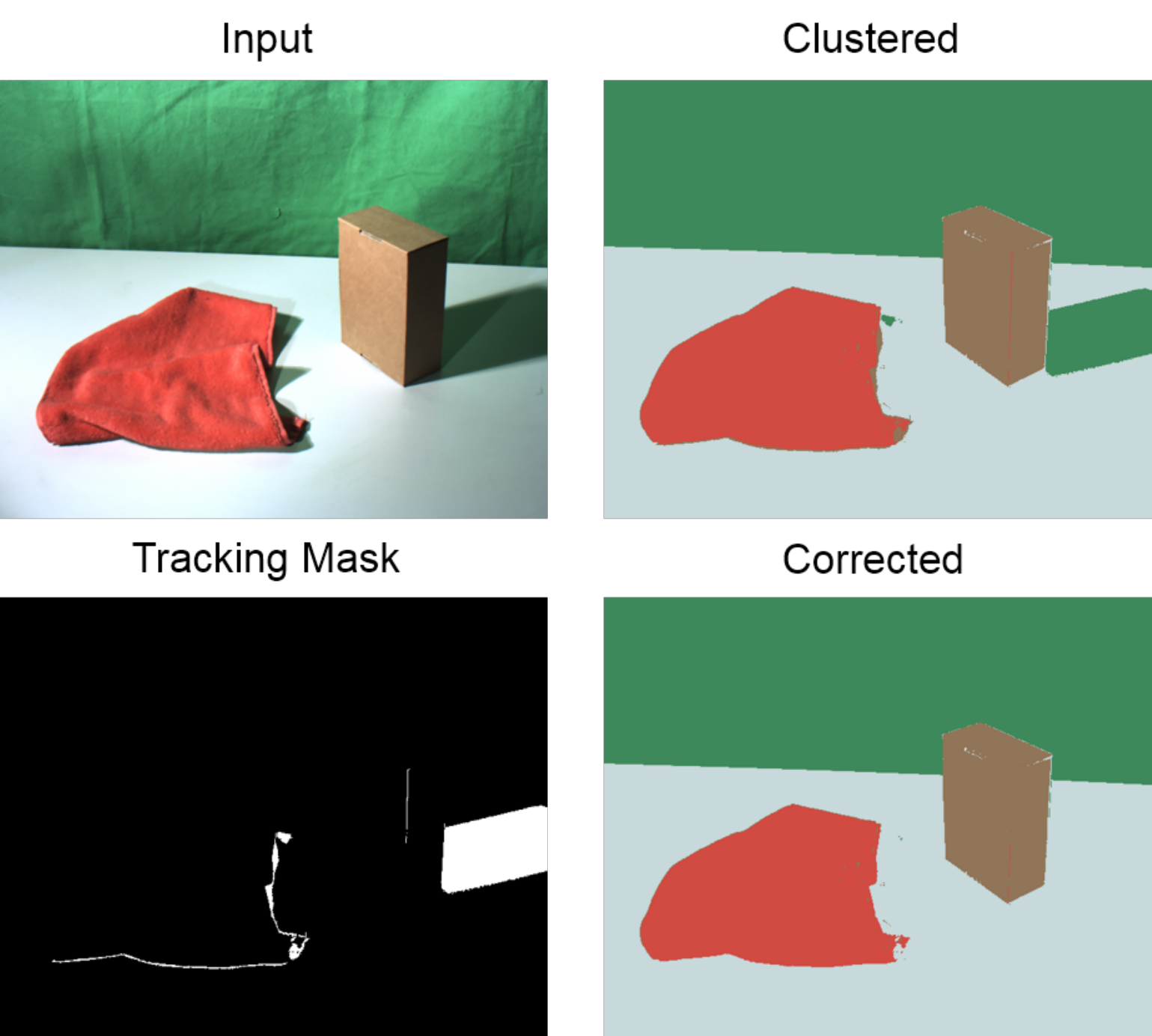}
	\caption{\label{fig:cluster-correction}%
		Example of misclustering correction (\cref{sec:misclustering}).
		The green color spill of the background causes misclustered regions in the shadow of the box and the towel (top right).
		We generate tracking masks (bottom left) using a few clicks for correcting the misclustered regions (bottom right).
	}
\end{figure}

\subsection{Misclustering Correction}
\label{sec:misclustering}

Since the clustering directly depends on the color of a pixel, regions of strong inter-reflections may be erroneously assigned to the base color of an indirect illuminant instead of the base color representing the reflectance of the region (see the green shadow of the box in \cref{fig:cluster-correction}).
Such a misclustering is difficult to correct automatically because of the inherent ambiguity of the illumination decomposition problem.
In this case, we rely on minimal manual interaction to identify misclustered regions and then automatically correct the underlying reflectance base color in all subsequent frames.

\subsubsection{Region Identification and Tracking}
Identifying the true reflectance of a pixel in the presence of strong inter-reflections from other objects is an ambiguous task. 
In case of direct illumination, the observed color value of a pixel is obtained by modulating the reflectance solely by the color of the illuminant (assumed to be white in our case).
However, in the case of inter-reflections, there is further modulation by light reflected from other objects, which then depends on their reflectance properties.
Such regions are easy to identify by a user, and so we ask the user to simply click on such a region \emph{only in the first frame} it occurs.
We then automatically identify the full region by flood filling it using connected-components analysis based on the cluster identifier.
In case the first fill does not cover the full region, additional clicks may be required.

\label{sec:tracking}

We use the following method for real-time tracking of non-rigidly deforming, non-convex marked regions in subsequent frames.
Given the marked pixel region in the previous frame, we probe the same pixel locations in the current frame to identify pixels with the same cluster ID as in the previous frame.
We flood fill starting from these valid pixels to obtain the tracked marked region in the new frame.
To keep this operation efficient, we do not flood fill for pixels inside the regions.
In practice, we observe that one or two valid pixels are sufficient to correctly identify the entire misclustered region.

\subsubsection{Reflectance Correction}

Once all pixels in a misclustered region are identified in a video frame (either marked or tracked), we exploit the sparsity constraint of the indirect illumination layers to solve for the correct reflectance base color.
We perform multiple full illumination decompositions (\cref{sec:GID}) for each identified region, evaluating each base color's suitability as the region's reflectance.
For each base color, we measure the sparsity obtained over the region using the illumination sparsity term to be introduced in \cref{eq:shading-sparsity-term}.
The base color that provides the sparsest solution of the decomposition is then used as the corrected reflectance.
The intuition behind such a sparsity prior is that using the correct underlying reflectance should lead to an illumination layer which is explained by the color spill from only a sparse number of nearby objects, as shown in \cref{fig:cluster-correction}.

\section{Illumination Decomposition}
\label{sec:energy}
\label{sec:GID}

Given the initial set of base colors for the scene, we next jointly decompose the input video and refine the base colors.
We decompose each input video frame $\mathbf{I}$ into its reflectance layer $\mathbf{R}$, its direct illumination layer $T_0$ and a set of indirect illumination layers $\{T_k\}$ corresponding to the base colors $\{\mathbf{b}_k\}$ (see \cref{sec:formulation}).
The decomposition into direct and multiple indirect illumination layers is inspired by \citet{CarroRA2011}.
The direct illumination layer~$T_0$ represents the direct contribution to the scene by the external light sources, and the indirect illumination layers $\{T_k\}$ capture the inter-reflections that occur within the scene.
We alternate this decomposition with the base color refinement (see \cref{sec:basecolorrefinement}).

We formulate our illumination decomposition as an energy minimization problem with the following energy:
\begin{equation}
\label{eq:giv-energy}
E_\mathrm{decomp}(\mathcal{X})
=
E_\mathrm{data}(\mathcal{X}) +
E_\mathrm{reflectance}(\mathcal{X}) +
E_\mathrm{illumination}(\mathcal{X})
\text{,}
\end{equation}
where $\mathcal{X} \!=\! \big\{\mathbf{R}, T_0, \{T_k\} \!\big\}$ is the set of variables to be optimized, while the base colors $\{\mathbf{b}_k\}$ stay fixed.
This energy has three main terms: the data fidelity term, reflectance priors (\cref{sec:reflectance-priors}) and illumination priors (\cref{sec:illumination-priors});
we give details on the individual energy terms below.
We optimize this energy using a novel fast GPU solver (see \cref{sec:optimization}) to obtain real-time performance.

\paragraph{Data Fidelity Term}
This constraint enforces that the decomposition result reproduces the input image:
\begin{equation}
\label{eq:data-term}
E_\mathrm{data}(\mathcal{X}) = \lambda_\mathrm{data} \cdot \sum_\mathbf{x}{
	\norm{\ 
		\mathbf{I}(\mathbf{x}) - \mathbf{R}(\mathbf{x}) \odot \sum_{k=0}^{K}{\mathbf{b}_k T_k(\mathbf{x})}
	\ }_2^2
} \text{,}
\end{equation}
where $\lambda_\mathrm{data}$ is the weight for this energy term (other terms have their own weights), and the $T_k$ are the $(K \!+\! 1)$ illumination layers of the decomposition: one direct layer $T_0$, and $K$ indirect layers~$\{T_k\}$.

\subsection{Reflectance Priors}
\label{sec:reflectance-priors}

We constrain the estimated reflectance layer $\mathbf{R}$ using three priors:
\begin{equation}
\label{eq:reflectance-energy}
E_\mathrm{reflectance}(\mathcal{X})
=
E_\mathrm{clustering}(\mathcal{X}) +
E_\mathrm{r\text{-}sparsity}(\mathcal{X}) +
E_\mathrm{r\text{-}consistency}(\mathcal{X})
\text{.}
\end{equation}
The first prior guides the illumination decomposition using the clustered chromaticity map of \cref{sec:clustering}, the second prior encourages a piecewise constant reflectance map using gradient sparsity, and the third prior is a global spatiotemporal consistency prior.

\paragraph{Reflectance Clustering Prior}

We use the clustering described in \cref{sec:clustering} to guide the decomposition, as the chroma\-ticity-clustered image $\mathbf{R}_\mathrm{cluster}$ is an approximation of the reflectance layer $\mathbf{R}$.
Hence, we constrain the reflectance map to remain close to the clustered image using the following energy term:
\begin{equation}
\label{eq:clustering-term}
E_\mathrm{clustering}(\mathcal{X})
= \lambda_\mathrm{clustering} \cdot \sum_\mathbf{x}{
	\norm{
		\mathbf{r}(\mathbf{x}) - \mathbf{r}_\mathrm{cluster}(\mathbf{x}) 
	}_2^2 \text{,}
}
\end{equation}
where the lowercase $\mathbf{r}$ represents the quantity $\mathbf{R}$ in the log-domain, i.e., $\mathbf{r} \!=\! \ln \mathbf{R}$,
and $\mathbf{r}_\mathrm{cluster}$ is the clustered reflectance map (\cref{sec:clustering}).

\paragraph{Reflectance Sparsity Prior}

Natural scenes generally consist of a small set of objects and materials, hence the reflectance layer is expected to have sparse gradients.
Such a spatially sparse solution for the reflectance image can be obtained by minimizing the $\ell_p$-norm ($p \!\in\! [0,1]$) of the gradient magnitude $\norm{\nabla \mathbf{r}}_2$.
Many intrinsic decomposition techniques \citep{MekaZRT2016,BonneSTSPP2014} have used similar reflectance sparsity priors:
\begin{equation}
\label{eq:reflectance-sparsity-term}
E_\mathrm{r\text{-}sparsity}(\mathcal{X})
= \lambda_\mathrm{r\text{-}sparsity} \cdot \sum_\mathbf{x}{
	\norm{\big. \nabla \mathbf{r}(\mathbf{x}) }_2^p
} \text{.}
\end{equation}

\paragraph{Spatiotemporal Reflectance Consistency Prior}

We also employ the spatiotemporal reflectance consistency prior $E_\mathrm{r\text{-}consistency}(\mathcal{X})$ first introduced by \citet{MekaZRT2016}.
This prior enforces that the reflectance stays temporally consistent by connecting every pixel with a set of randomly sampled pixels in a small spatiotemporal window by constraining the reflectance of the pixels to be close under a defined chromaticity-closeness condition.
We refer to \citet{MekaZRT2016} for further details.

\subsection{Illumination Priors}
\label{sec:illumination-priors}

We constrain the illumination $\mathbf{S}$ to be close to monochrome and the indirect illumination layers $\{T_k\}$ to have a sparse decomposition, spatial smoothness and non-negativity:
\begin{align}
\label{eq:shading-energy}
\begin{split}
E_\mathrm{illumination}(\mathcal{X})
&=
E_\mathrm{monochrome}(\mathcal{X}) +
E_\mathrm{i\text{-}sparsity}(\mathcal{X}) \\
&+ E_\mathrm{smoothness}(\mathcal{X}) +
E_\mathrm{non\text{-}neg}(\mathcal{X})
\text{.}
\end{split}
\end{align}

\paragraph{Soft-Retinex Weighted Monochromaticity Prior}
The illumination layer is a combination of direct and indirect illumination effects.
Indirect effects such as inter-reflections tend to be spatially local with smooth color gradients whereas under the white-illumination assumption, the direct bounce does not contribute any color to the illumination layer.
Hence, we expect the illumination $\mathbf{S}$ to be mostly monochromatic except at small spatial pockets where smooth color gradients occur due to inter-reflections.
Therefore, we impose the following constraint:
\begin{equation}
E_\mathrm{monochrome}(\mathcal{X})
=
\lambda_\mathrm{monochrome} \cdot w_\mathrm{SR} \cdot \sum_{\mathbf{x}} \sum_c (\mathbf{S}_c(\mathbf{x}) - \abs{\mathbf{S}(\mathbf{x})})^2
\text{,}
\end{equation}
where $c \!\in\! \{R, \!G, \!B\}$, and $\abs{\mathbf{S}}$ is the intensity of the illumination layer~$\mathbf{S}$.
This constraint pulls the color channels of each pixel close to the grayscale intensity of the pixel, hence encouraging monochromaticity.
$w_\mathrm{SR}$ is the soft-color-Retinex weight computed using
\begin{equation}
w_\mathrm{SR} =  1 - \exp(-50 \cdot \Delta \mathbf{C}) \text{.}
\end{equation}
Here, $\Delta \mathbf{C}$ is the maximum of the chromaticity gradient of the input image in any of the four spatial directions at the pixel location.
The soft-color-Retinex weight is high only for large chromaticity gradients, which represent reflectance edges.
Hence, monochromaticity of the illumination layer is enforced only close to the reflectance edges and not at locations of slowly varying chromaticity, which represent inter-reflections.
Relying on local chromaticity gradients may be problematic when there are regions of uniform colored reflectance, but in such regions the reflectance sparsity priors tend to be stronger and overrule the monochromaticity prior.

\paragraph{Illumination Decomposition Sparsity}
We enforce that the illumination decomposition is sparse in terms of the layers that are activated per-pixel, i.e., those that influence the pixel with their corresponding base color.
Here, the assumption is that during image formation in the real world, a large part of the observed radiance for a scene point comes from a small subpart of the scene.
To achieve decomposition sparsity, we follow \citet{CarroRA2011} and apply the sparsity-inducing $\ell_1$-norm \citep{BachJMO2012} to the indirect illumination layers:
\begin{equation}
\label{eq:shading-sparsity-term}
E_\mathrm{i\text{-}sparsity}(\mathcal{X})
=
\lambda_\mathrm{i\text{-}sparsity} \cdot \sum_\mathbf{x}{
	\norm{ \big\{ T_k(\mathbf{x}) \big\}_{k=1}^K }_1
}
\text{.}
\end{equation}

\paragraph{Spatial Smoothness}
We further encourage the decomposition to be spatially piecewise smooth using an $\ell_1$-sparsity prior in the gradient domain, similar to \citet{CarroRA2011}, which enforces piecewise constancy of each direct or indirect illumination layer:
\begin{equation}
\label{eq:smoothness-term}
E_\mathrm{smoothness}(\mathcal{X})
=
\lambda_\mathrm{smoothness} \cdot
\sum_\mathbf{x}{
	\sum_{k=0}^{K}{
		\norm{\big.\nabla T_k(\mathbf{x})}_1 }
}
\text{.}
\end{equation}
This allows to have sharp edges in the decomposition layers.

\paragraph{Non-Negativity of Light Transport}
Light transport is an inherently additive process: light bouncing around in the scene adds radiance
to scene points, but never subtracts from them.
Thus, the quantity of transported light is always positive.
Since our illumination decomposition layers are motivated by physical light transport, we enforce them to be non-negative to obey this principle:
\begin{equation}
\label{eq:nonnegativityterm}
E_\mathrm{non\text{-}neg}(\mathcal{X})
=
\lambda_\mathrm{non\text{-}neg} \cdot
\sum_\mathbf{x}{
	\sum_{k=0}^{K}{
		\max\left(-T_k(\mathbf{x}), 0\right)
	}
} \text{.}
\end{equation}
If the decomposition layer $T_k(\mathbf{x})$ is non-negative, there is no penalty.
Otherwise, if $T_k(\mathbf{x})$ becomes negative, a linear penalty is enforced.

\begin{figure}
	\includegraphics[width=\linewidth]{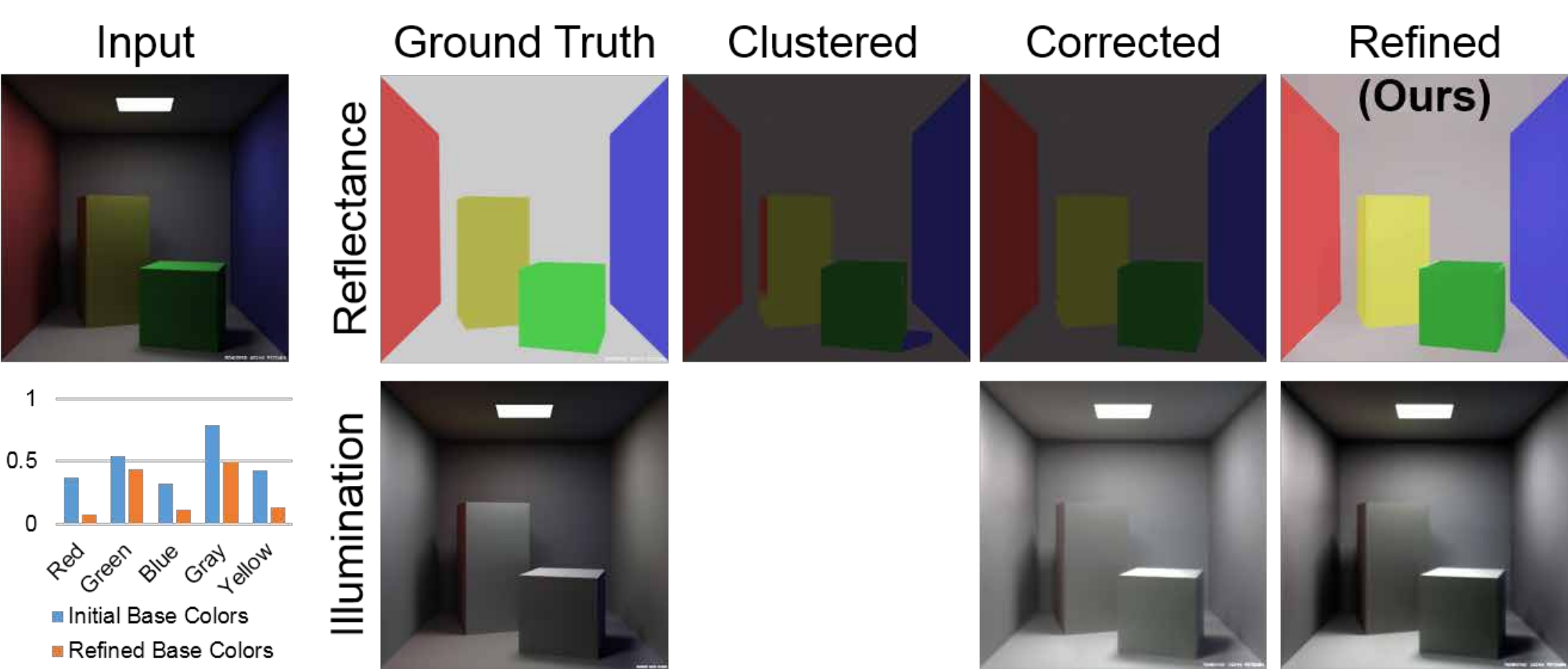}
	\caption{\label{fig:basecolorrefinement}%
		Here we show the improvement obtained by the base color refinement in our approach.
		We start from the clustered reflectance map (Clustered).
		Our misclustering correction leads to a better segmentation of the scene (Corrected).
		Finally, our approach fully automatically optimizes for a better set of base colors (Refined).
		As can be seen, our base color refinement leads to a significant improvement and results closer to the ground truth.
		The bar chart shows the error between the ground-truth base colors and our estimated base color with (orange) and without (blue) base color refinement.
	}
\end{figure}

\subsection{Base Color Refinement}
\label{sec:basecolorrefinement}

We estimate the initial base colors using chromaticity-based histogram clustering (\cref{sec:clustering}).
Unlike previous methods that keep the base colors fixed once estimated \citep{CarroRA2011,AksoyAPS2016}, we refine the base colors further on the first video frame to approach the ground-truth reflectance of the materials in the scene.
The refinement of base colors is formulated as an incremental update $\Delta\mathbf{b}_k$ of the base colors $\mathbf{b}_k$ in the original data fidelity term (\cref{eq:data-term}), along with intensity and chromaticity regularizers:
\begin{align}
E_\mathrm{refine}(\mathcal{X})
&=
\lambda_\mathrm{data} \sum_\mathbf{x}	\norm{\ 
	\mathbf{I}(\mathbf{x}) - \mathbf{R}(\mathbf{x}) \odot \sum_{k=0}^{K}{(\mathbf{b}_k \!+\! \Delta\mathbf{b}_k) T_k(\mathbf{x})}
\ }_2^2 \label{eq:basecolorrefinement}\\
&+ \lambda_\mathrm{IR} \sum_{k=1}^K \norm{\big. \Delta\mathbf{b}_k}_2^2
+ \lambda_\mathrm{CR} \sum_{k=1}^K \norm{\big. (\mathbf{C}(\mathbf{b}_k) + \Delta\mathbf{b}_k) - \mathbf{C}(\mathbf{b}_k)}_2^2
\text{.} \nonumber
\end{align}

Here, $\mathcal{X} \!=\! \{\Delta\mathbf{b}_k\}$ is the vector of unknowns to be optimized,
$\lambda_\mathrm{IR}$ is the weight for the intensity regularizer that ensures small base color updates, and 
$\lambda_\mathrm{CR}$ is the weight of the chromaticity regularizer, which constrains base color updates $\Delta\mathbf{b}_k$ to remain close in chromaticity $\mathbf{C}(\cdot)$ to the initially estimated base color $\mathbf{b}_k$.
These regularizers ensure that base color updates do not lead to oscillations in the optimization process.
The refinement energy is solved in combination with the illumination decomposition energy (\cref{eq:giv-energy}), resulting in an estimation of the unknown variables that together promotes decomposition sparsity.
See \cref{fig:basecolorrefinement} for an example.

This refinement of the base colors leads to a dense Jacobian matrix, because the unknown variables $\{\Delta\mathbf{b}_k\}$ in the energy are influenced by all pixels in the image.
This makes the resulting optimization problem difficult to solve in a parallel fashion.
We present our solution to this issue in \cref{sec:optimization}.

\subsection{Handling the Sparsity-Inducing Norms}

Some energy terms contain sparsity-inducing $\ell_p$-norms ($p \!\in\! [0,1]$), i.e., \cref{eq:reflectance-sparsity-term,eq:shading-sparsity-term,eq:smoothness-term}.
We handle these objectives in a unified manner using Iteratively Re-weighted Least Squares \cite{HollaW1977}.
To this end, we approximate the $\ell_p$-norms by a non-linear least-squares objective based on re-weighting, i.e., we replace the corresponding residuals $\mathbf{r}$ as follows:
\begin{align}
	\norm{\mathbf{r}}_p
	&= \norm{\mathbf{r}}_2^2 \cdot \norm{\mathbf{r}}_2^{p-2} \\
	&\approx \norm{\mathbf{r}}_2^2 \cdot \underbrace{\norm{\mathbf{r}_{\textrm{old}}}_2^{p-2}}_{\text{constant}}
\end{align}
in each step of the applied iterative solver, see also \cref{sec:optimization}.
Here, $\mathbf{r}_\textrm{old}$ is the corresponding residual after the previous iteration step.

\subsubsection{Handling Non-negativity Constraints}

The non-negativity objective in \cref{eq:nonnegativityterm} contains a maximum function that is non-differentiable at zero.
As proposed by \citet{CarroRA2011}, we handle this objective by replacing the maximum with a re-weighted least-squares term, $\max(-T_k(\mathbf{x}), 0) = w_k T_k^2(\mathbf{x})$, using
\begin{equation}
w_k =
\begin{cases}
0 & \text{if } T_k(\mathbf{x}) > 0 \\
\left(\abs{T_k(\mathbf{x})} + \epsilon\right)^{-1} & \text{otherwise} \text{.}
\end{cases}
\end{equation}
Here, $\epsilon=0.002$ is a small constant that prevents division by zero.
This transforms our non-convex energy into a non-linear least-squares optimization problem.

\section{Data-Parallel GPU Optimization}
\label{sec:optimization}

Our decomposition problems are all non-convex optimizations based on an objective $E$ with unknowns $\mathcal{X}$.
We find the best decomposition $\mathcal{X}^{*}$ by solving the following minimization problem:
\begin{equation}
\label{eq:optimization}
\mathcal{X}^{*} = \argmin_{\mathcal{X}}{E}(\mathcal{X}) \text{.}
\end{equation}
The optimization problems are in general non-linear least-squares form and can be tackled by the iterative Gauss--Newton algorithm that approximates the optimum $\mathcal{X}^* \!\approx\! \mathcal{X}_k$ by a sequence of solutions $\mathcal{X}_k \!=\! \mathcal{X}_{k-1} + \boldsymbol \delta_{k}^*$.
The optimal linear update $\boldsymbol\delta_{k}^*$ is given by the solution of the associated normal equations:
\begin{equation}
\boldsymbol\delta_{k}^{*} = \argmin_{\boldsymbol\delta_{k}}{ \norm{ \ \big. \mathbf{F}(\mathcal{X}_{\mathrm{k-1}}) + \boldsymbol\delta_{k} \, \mathbf{J}(\mathcal{X}_{\mathrm{k-1}}) \ } }_2^2 \text{.}
\end{equation}
Here, $\mathbf{F}$ is a vector field that stacks all residuals, i.e., $E(\mathcal{X}) \!=\! \norm{ \mathbf{F}(\mathcal{X}) }_2^2$, and $\mathbf{J}$ is its Jacobian matrix.

Obtaining real-time performance is challenging even with recent state-of-the-art data-parallel iterative non-linear least-squares solution strategies \cite{MekaZRT2016,ZollhNIRZFWFLTS2014,WuZNSIT2014}.
To see why this is the case, let us have a closer look at the normal equations.
To avoid cluttered notation, we will omit the parameters and simply write $\mathbf{J}$ instead of $\mathbf{J}(\mathcal{X})$.
For our decomposition energies, the Jacobian $\mathbf{J}$ is a large matrix with usually more than 70 million rows and 4 million columns.
Previous approaches assume $\mathbf{J}$ to be a sparse matrix, meaning that only a few residuals are influenced by each variable.
While this holds for the columns of $\mathbf{J}$ that corresponds to the variables that are associated with the decomposition layers, it does not hold for the columns that store the derivatives with respect to the base color updates $\{\Delta\mathbf{b}_k\}$, since the base colors influence each residual of $E_\mathrm{data}$ (\cref{eq:data-term}).
Therefore,
$
\mathbf{J} \!=\! \left[\mathbf{S_J} \ \ \mathbf{D_J} \right]
$
has two sub-blocks:
$\mathbf{S_J}$ is a large sparse matrix with only a few non-zero entries per row, while $\mathbf{D_J}$ is dense, with the same number of rows, but only a few columns.
Thus, the evaluation of the Jacobian $\mathbf{J}$ requires a different specialized parallelization for the dense and sparse parts.

\subsection{Sparse–Dense Splitting}

We tackle the described problem using a sparse–dense splitting approach that splits the variables $\mathcal{X}$ into a sparse set $\mathcal{T}$ (decomposition layers) and a dense set $\mathcal{B}$ (base color updates).
Afterwards, we optimize for $\mathcal{B}$ and $\mathcal{T}$ independently in an iterative flip-flop manner.
First, we optimize for $\mathcal{T}$, while keeping $\mathcal{B}$ fixed.
The resulting optimization problem is a sparse non-linear least-squares problem.
Thus, we improve upon the previous solution by performing a non-linear Gauss--Newton step.
The corresponding normal equations are solved using 16 steps of data-parallel preconditioned conjugate gradient.
We parallelize over the rows of the system matrix using one thread per row (variable).

After updating the `sparse' variables $\mathcal{T}\!\!$, we keep them fixed and solve for the `dense' variables $\mathcal{B}$.
The resulting optimization problem is a dense least-squares problem with a small $3K \times 3K$ system matrix (normally $K$ is between 4 and 7 due to merged clusters).
We materialize the normal equations in device memory based on a sequence of outer products, using one thread per entry of $\mathbf{J}^\top \mathbf{J}$.
Finally, the system is mapped to the CPU and robustly solved using singular value decomposition.
After updating `dense' variables $\mathcal{B}$, we again solve for `sparse' variables $\mathcal{T}$ and iterate this process until convergence.


\section{Results and Evaluation}
\label{sec:results}

\begin{figure}
	\includegraphics[width=\linewidth]{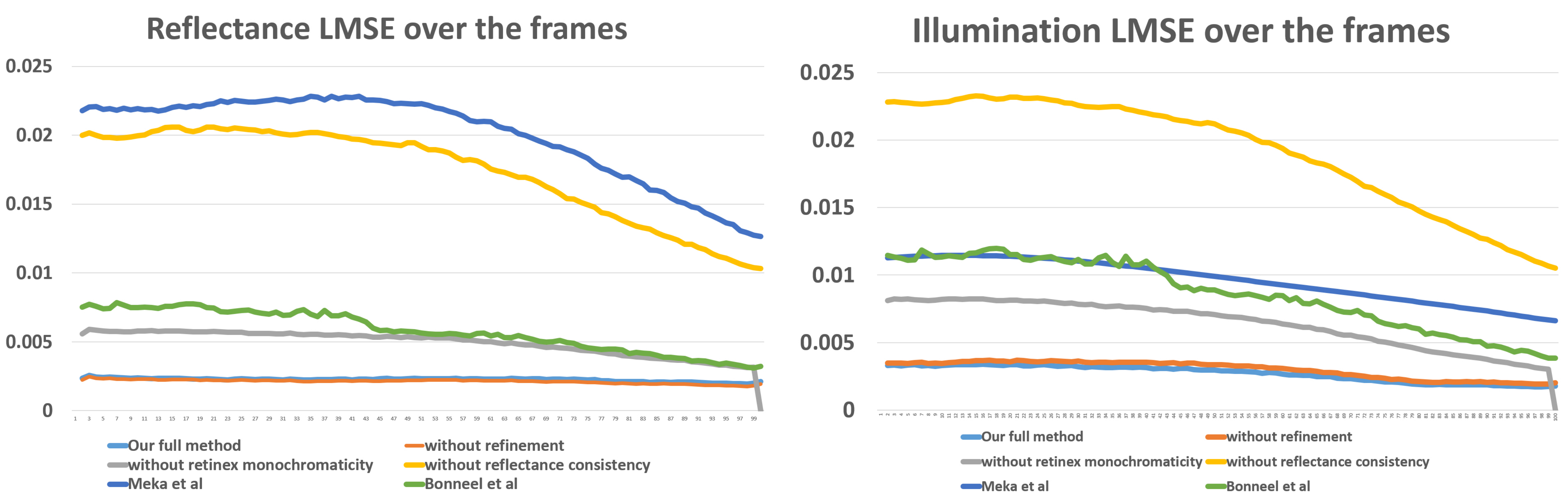}
	\caption{\label{fig:BaseColorRefinementGraphs}%
		We quantitatively analyze our method on the \textsc{SyntheticRoom} sequence.
		We plot the LMSE error \citep{GrossJAF2009} per frame in this graph.
		Our full method achieves the lowest average LMSE of 0.002.
		Without the base color refinement, the average error is 0.0025 and results look visibly worse.
		Without the Retinex-monochromaticity and the reflectance consistency priors, the average errors are even higher: 0.005 and 0.019, respectively.
		We also compare two other decomposition techniques: \citet{MekaZRT2016} has an average error of 0.015, and \citet{BonneSTSPP2014} has 0.007.
	}
\end{figure}

We now show results obtained with our approach, evaluate them qualitatively and quantitatively, and compare to current state-of-the-art decomposition approaches.
Please note that we scale the indirect illumination layers for better visualization.
We performed our evaluation in terms of robustness, accuracy and runtime on a dataset containing several challenging real and synthetic video sequences.
The used test datasets consists of fourteen real and one synthetic sequence
(\textsc{Boat}, \textsc{Box}, \textsc{Box2}, \textsc{Cart}, \textsc{ChitChat}, \textsc{Cups}, \textsc{Droid}, \textsc{Girl}, \textsc{Girl2}, \textsc{Hands}, \textsc{Kermit}, \textsc{Toys}, \textsc{Umbrella} and \textsc{SyntheticRoom}).
We refer to the accompanying video for the results on the complete video sequences.
We compare
to the intrinsic decomposition approaches of \citet{BonneSTSPP2014} and \citet{MekaZRT2016}, and the illumination decomposition approach of \citet{CarroRA2011}.
Our approach is much faster than previous decomposition techniques, and it obtains higher-quality decomposition results in terms of the reflectance map and the indirect illumination layers, which directly translates to higher-quality results in all shown applications.

\paragraph{Parameters}

We used the following fixed set of parameters in all our experiments:
$\lambda_\mathrm{clustering} \!=\! 200$,
$\lambda_\mathrm{r\text{-}sparsity} \!=\! 20$,
$p \!=\! 1$,
$\lambda_\mathrm{i\text{-}sparsity} \!=\! 3$,
$\lambda_\mathrm{smoothness} \!=\! 3$,
$\lambda_\mathrm{non\text{-}neg} \!=\! 1000$,
$\lambda_\mathrm{data} \!=\! 5000$, 
$\lambda_\mathrm{IR} \!=\! 10$,
$\lambda_\mathrm{CR} \!=\! 100$ and
$\lambda_\mathrm{r\text{-}consistency} \!=\! \lambda_\mathrm{monochrome} \!=\! 10$.
Since $\lambda_\mathrm{data}$ is set to a high value, the residual of the data term (\cref{eq:data-term}) is below one percent of the intensity range; hence it is too dark to see.

\paragraph{Runtime Performance}

We measured the performance of our approach on an Intel Core i7 with 2.7\,GHz, 32\,GB RAM and an NVIDIA GeForce GTX 980.
The runtime for videos with a resolution of 640$\times$512\,pixels can be broken down into:
14\,ms for illumination decomposition,
2\,s for base color refinement, and
1\,s for misclustering correction.
Note that we perform the last two steps, base color refinement and misclustering correction, only once at the beginning of the video.
Afterwards, our approach runs at real-time frame rates ($\geqslant$30\,Hz) and enables real-time video editing applications.

\subsection{Quantitative Results}

We perform quantitative evaluation on our \textsc{SyntheticRoom} sequence.
The sequence was rendered using Blender's Cycles renderer.
All objects in the scene are assigned diffuse materials, with natural white illumination from the window.
The objects in the scene cause significant inter-reflections.
We also render the ground-truth reflectance and illumination images.
We compare our decomposition to the ground-truth sequences and compute the LMSE error metric proposed by \citet{GrossJAF2009}.
We plot the LMSE error per-frame in \Cref{fig:BaseColorRefinementGraphs}. 
We perform an ablation study by analyzing the error with our full energy and then removing some of the energy components, such as the base color refinement, the Retinex-monochromaticity prior and the spatiotemporal reflectance consistency prior.
We also compare against state-of-the-art intrinsic video decomposition techniques. 
Our full method obtains the best results.

\begin{figure}
	\includegraphics[width=\linewidth]{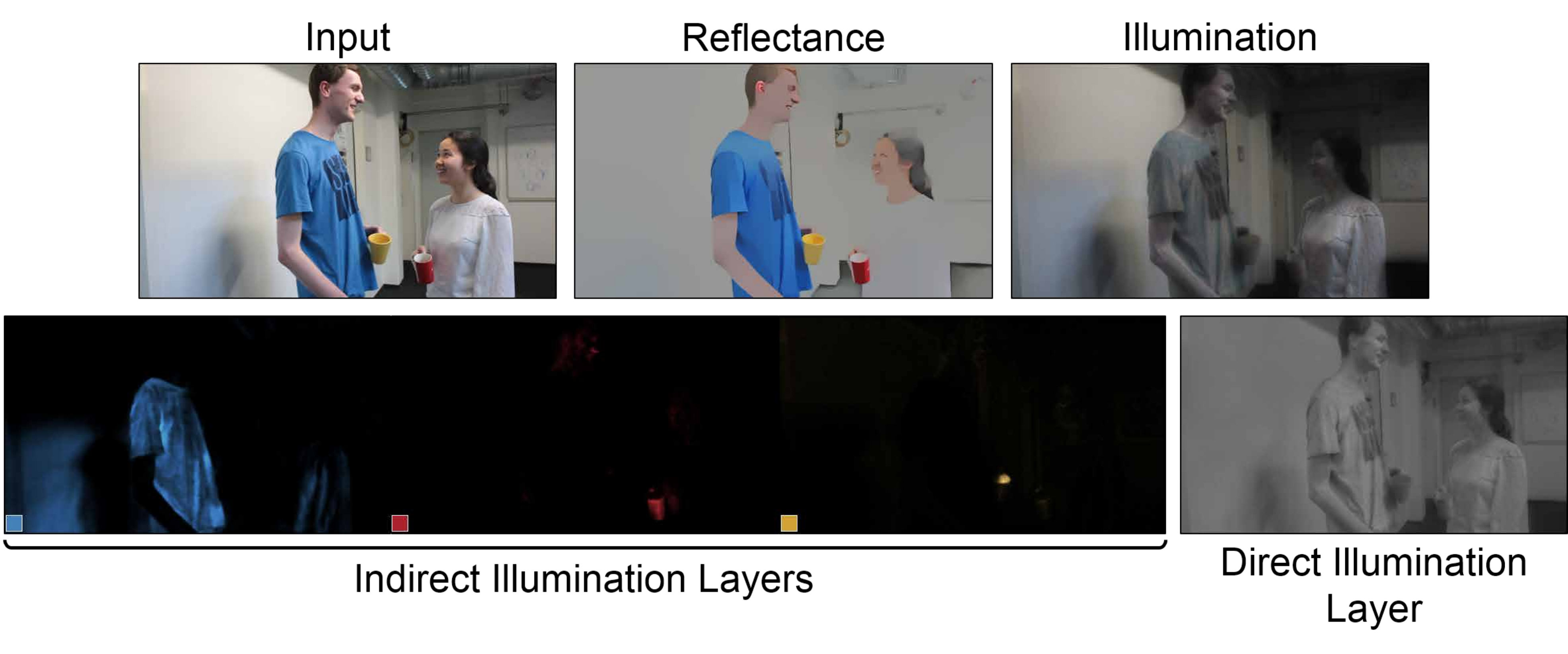}
	\caption{\label{fig:chitchat1}%
		Our decomposition of the \textsc{ChitChat} sequence.
		We accurately decompose the color spill from the blue shirt and the red cup.
		Note that the reflectance is devoid of both color spills.
	}
\end{figure}

\begin{figure}
	\includegraphics[width=\linewidth]{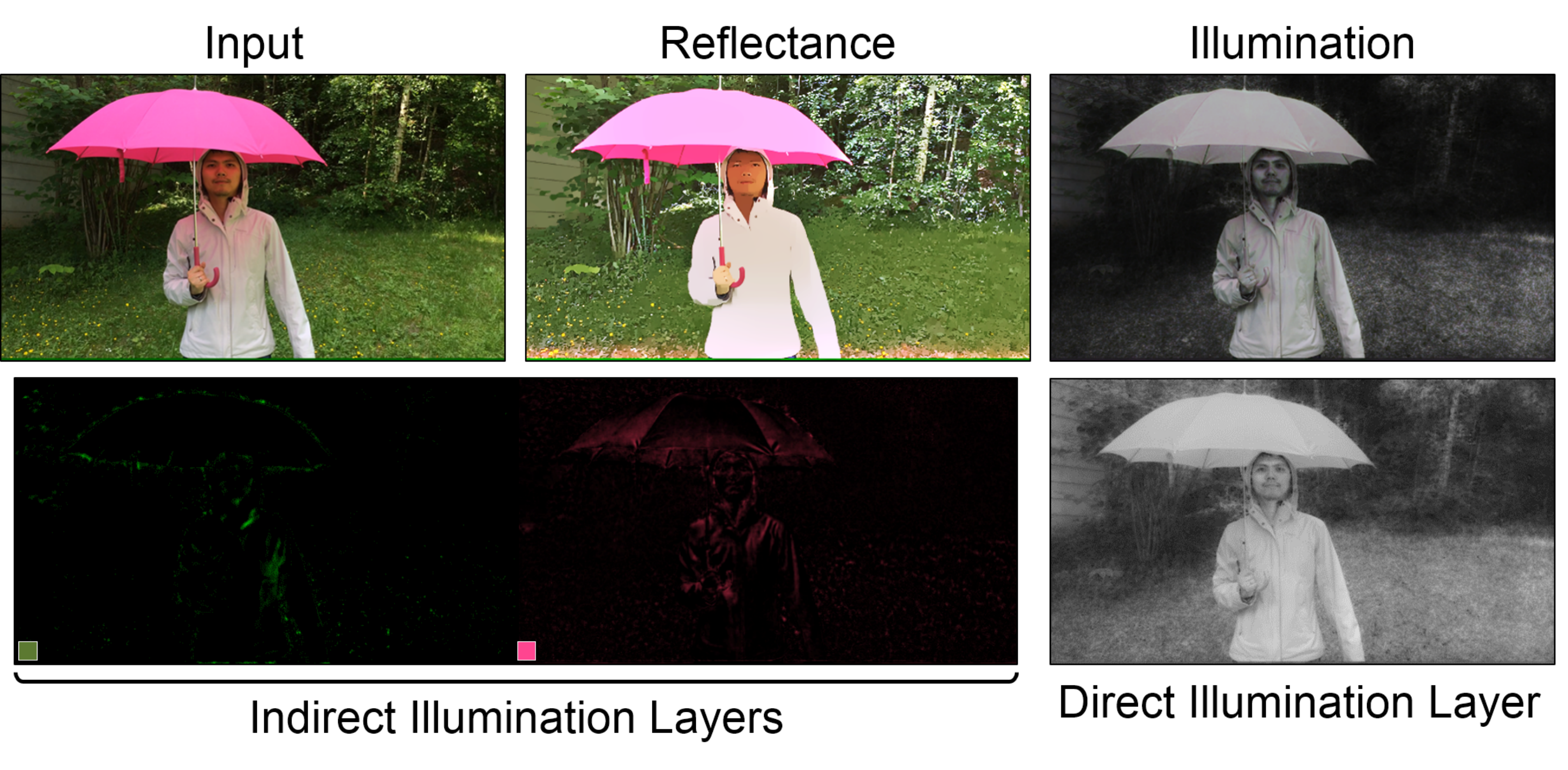}
	\caption{\label{fig:umbrella}%
		Our decomposition of the \textsc{Umbrella} sequence.
		The complex color spill from the umbrella is mixed with the spill from the forest on the face and the jacket.
		Our method is able to decompose the colors accurately.
		Note that the reflectance is free from either of the two color spills, and that both are present in the respective indirect illumination layers.
	}
\end{figure}

\subsection{Qualitative Results}

\begin{figure}
	\includegraphics[width=\linewidth]{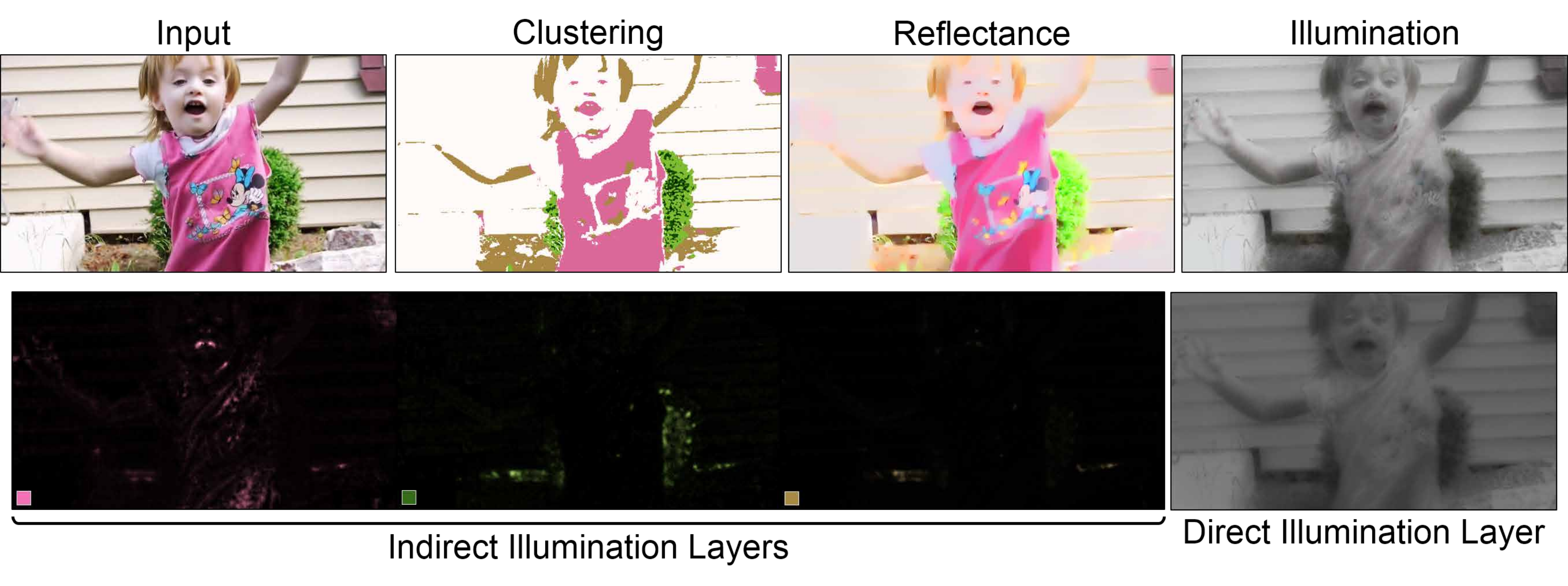}
	\caption{\label{fig:girl2}%
		Our decomposition of the \textsc{Girl2} sequence.
		Even in challenging scenes, where the color palette is not well defined and thus clustering is difficult, our approach is able to estimate a plausible  decomposition along with various indirect illumination layers.
		Note the strong pink inter-reflection on the neck of the girl and within the shirt and in the green bush.
	}
\end{figure}

\begin{figure}
	\includegraphics[width=\linewidth]{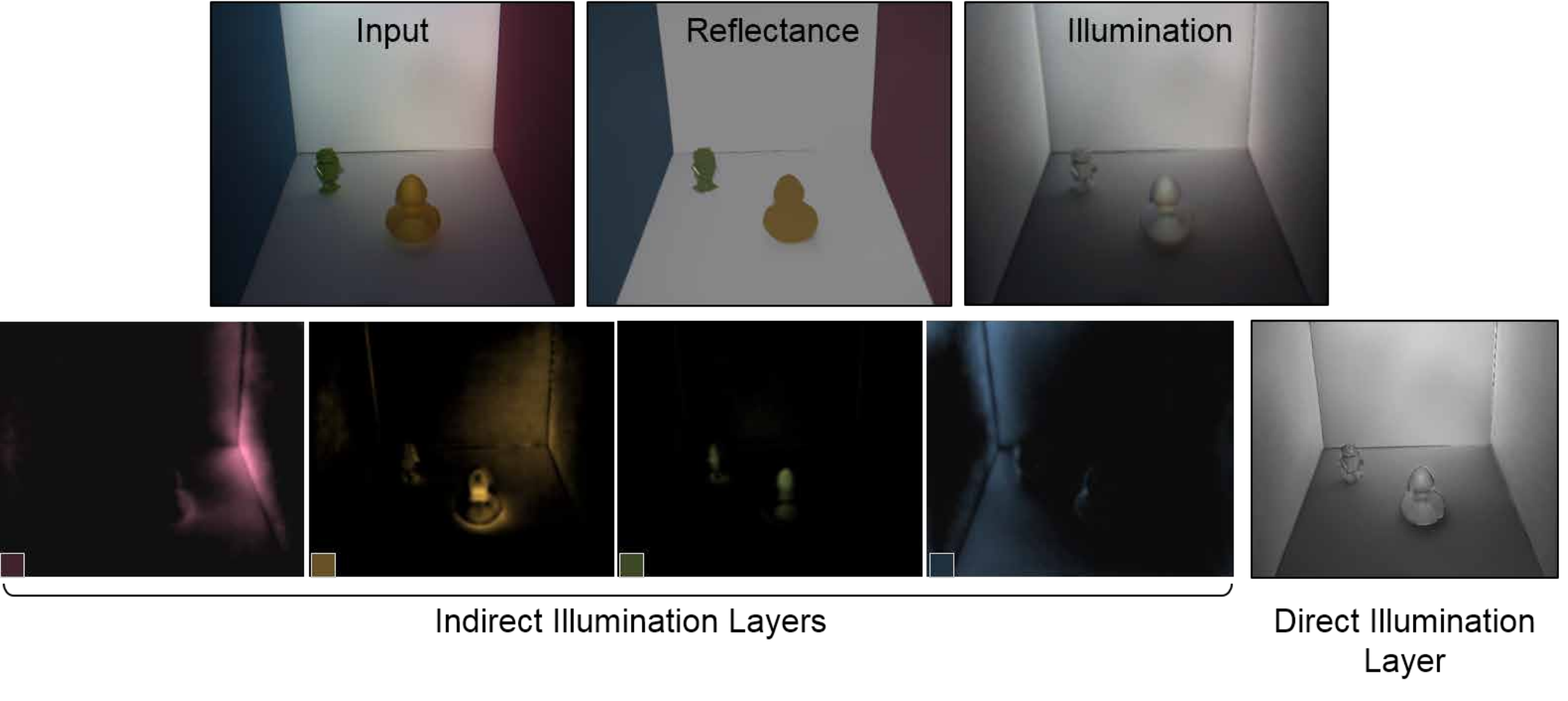}
	\caption{\label{fig:cornell1}%
		Our decomposition of the \textsc{Droid} sequence.
		Note the clean reflec\-tance map and clearly separated color casts in the indirect illumination layers.
	}
\end{figure}

We show that the indirect illumination layers computed by our approach at real-time frame rates nicely capture the inter-reflections between various kinds of objects in a consistent manner, see \cref{fig:teaser,fig:chitchat1,fig:umbrella,fig:cornell1,fig:girl2}.
The brightness for all the indirect illumination layers shown in the paper has been scaled by $2\times$ for better visualization.
In contrast to intrinsic decomposition approaches, ours separates the input image into reflectance, colored direct and indirect illumination layers.
Please note the color bleeding of the different parts of the boat in \cref{fig:teaser}, which is clearly visible and nicely reconstructed, even though it only accounts for a small amount of the lighting in the input image.

\Cref{fig:girl2} shows the illumination decomposition for a complex scene with fast motion and a difficult color palette. 
Our clustering strategy fails to achieve a meaningful segmentation of the scene.
Yet, we are able to produce a plausible decomposition of this challenging scene.
In particular, we are able to capture the color spill from the girl's shirt to her neck and the inter-reflections on the ground from the bush in the background.

\begin{figure}
	\includegraphics[width=\linewidth]{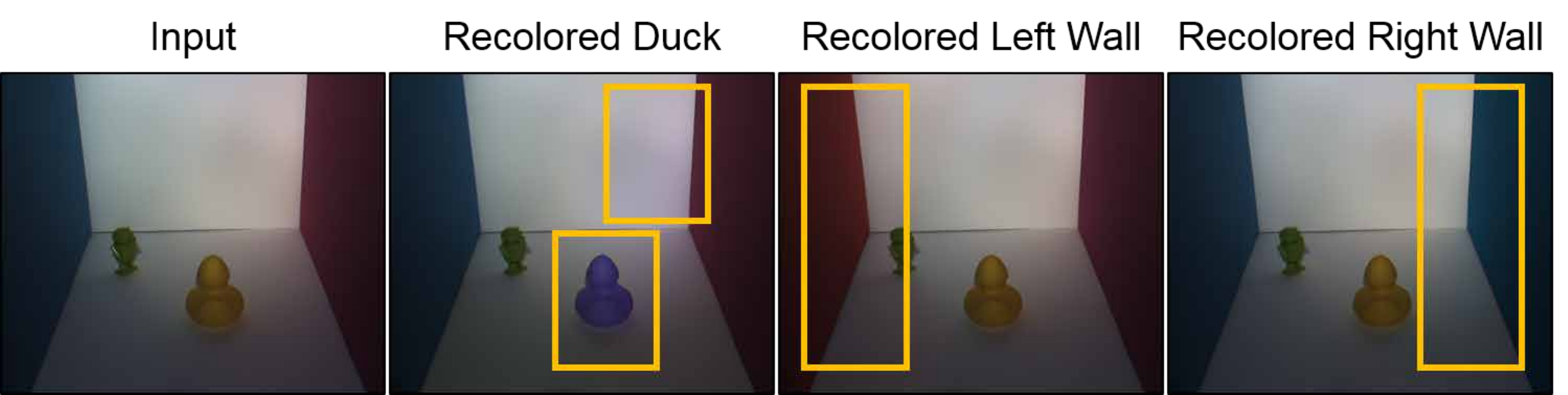}
	\caption{\label{fig:cornell2}%
		Our approach enables live recoloring of scene surfaces in a photo-realistic and globally consistent manner.
		Here, we recolor the rubber duck and the walls in the scene.
		Note how the corresponding global illumination in the scene (highlighted) is also consistently modified by our approach.
	}
\end{figure}

\begin{table}\small
	\caption{\label{tab:number-of-interactions}%
		User interactions required for all sequences shown in this paper and our supplemental video.
		Note that most sequences do not require any user interaction (bottom half of the table).
	}
	\begin{tabular}{l@{\hspace{16pt}}c@{\hspace{16pt}}r}
		\textbf{Sequence}                   &                      \textbf{Figures}                      & \textbf{Interactions} \\ \hline
		\textsc{\textsf{Box2}}              &                                                            &                     2 \\
		\textsc{\textsf{Box}}               &        \ref{fig:cluster-correction}, \ref{fig:box}         &                     3 \\
		\textsc{\textsf{Cart}} &                       \ref{fig:cart}                       &                    1 \\
		\textsc{\textsf{Cups}}              &        \ref{fig:mug}, \ref{fig:software_comparison}        &                     1 \\
		\textsc{\textsf{Hands}}             &                                                            &                     1 \\
		\textsc{\textsf{Toys}}              &             \ref{fig:pipeline}, \ref{fig:toy}              &                     5 \\ \hline
		\textsc{\textsf{Boat}}              &                      \ref{fig:teaser}                      &                     0 \\
		\textsc{\textsf{ChitChat}}          &          \ref{fig:chitchat1}, \ref{fig:chitchat2}          &                     0 \\
		\textsc{\textsf{Cornell}}           &     \ref{fig:basecolorrefinement}, \ref{fig:synthetic}     &                     0 \\
		\textsc{\textsf{Droid}}             & \ref{fig:cornell1}, \ref{fig:cornell2}, \ref{fig:sparsity} &                     0 \\
		\textsc{\textsf{Girl}}              &                                                            &                     0 \\
		\textsc{\textsf{Girl2}}             &                      \ref{fig:girl2}                       &                     0 \\
		\textsc{\textsf{Kermit}}            &                     \ref{fig:kermit1}                      &                     0 \\
		\textsc{\textsf{Paper}}             &                      \ref{fig:paper}                       &                     0 \\
		\textsc{\textsf{Umbrella}}          &                     \ref{fig:umbrella}                     &                     0 \\
		\textsc{\textsf{SyntheticRoom}}     &                    \ref{fig:synthetic2}                    &                     0
	\end{tabular} 
\end{table}

\Cref{fig:cornell1} shows another example of the reconstructed illumination layers, where the color bleeding of the red and blue walls onto the floor is clearly visible.
This sequence also shows that our decomposition is temporally coherent and that the illumination layers instantly adapt to changes in the scene.
This can best be seen in the supplemental video.
Such a decomposition into direct and indirect illumination is of paramount importance for illumination-consistent recoloring.
We show an example of this for the same scene in \cref{fig:cornell2}.
Here, we first recolor the yellow duck to purple, which influences the color of the floor.
In another example, we recolor the walls from blue to red, and vice versa, which also consistently changes the inter-reflections on the floor.
Please note that our decomposition is computed at real-time frame rates, which enables the user to explore these effects interactively.
In \cref{tab:number-of-interactions}, we list the number of user-clicks that were performed for each sequence.
Please note that most of the sequences did not require user interaction.
Where necessary, we required only a small number of clicks, owing to our region-tracking strategy.
\begin{figure*}
	\centering
	\input{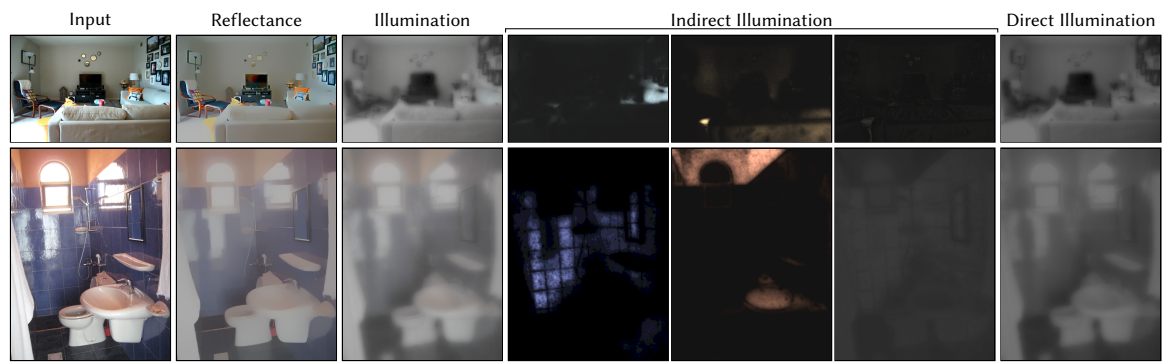}
	\caption{\label{fig:IIW}%
		Our illumination decomposition applied to two samples from the ‘Intrinsic Images in the Wild’ dataset \citep{BellBS2014}.
	}
\end{figure*}

To evaluate our method on more general and more complex scenes, which consist of more than just a few prominent objects, we test our method on images from the \textit{Intrinsic Images in the Wild} dataset \citep{BellBS2014}.
This dataset consists of room-sized indoor scenes.
Even though such scenes generally do not exhibit particularly strong global lighting effects, our method is still able to pick up the prominent colors and visualize the global color spills that occur due to them, as shown in \cref{fig:IIW}.
Such scenes are challenging for our method to handle, but video editing tasks such as recoloring can still benefit from our decomposition, even in such a challenging setting.
We obtain a weighted human disagreement rate (WHDR) of 27.2\%, which is better than the baselines and other video decomposition techniques such as \citet{MekaZRT2016}.

\begin{figure*}
	\includegraphics[width=\linewidth]{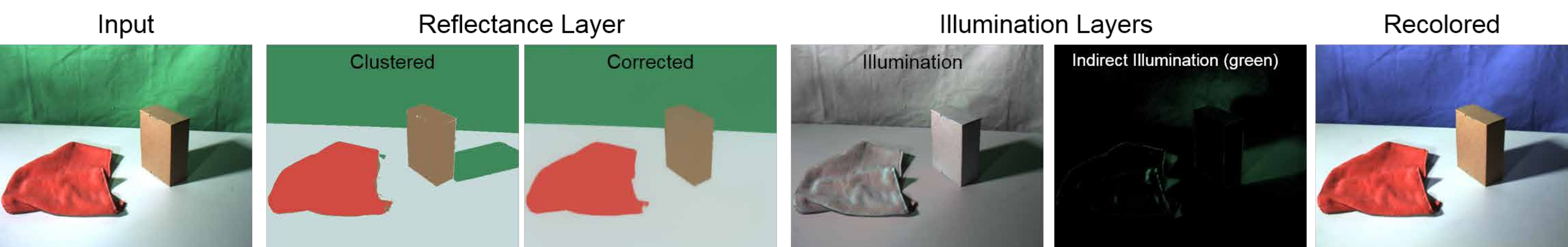}
	\caption{\label{fig:box}%
		Results on the \textsc{Box} sequence, with and without our novel sparsity-based misclustering correction.
		Regions with strong inter-reflections (shadow of the box) are often misclustered in the reflectance image.
		This causes indirect illumination to wrongly influence the reflectance layer and not the illumination layer, which makes inter-reflection-consistent recoloring impossible.
		Our method alleviates this problem with a little bit of user input to correct the misclustering.
	}
\end{figure*}

\begin{figure}
	\includegraphics[width=\linewidth]{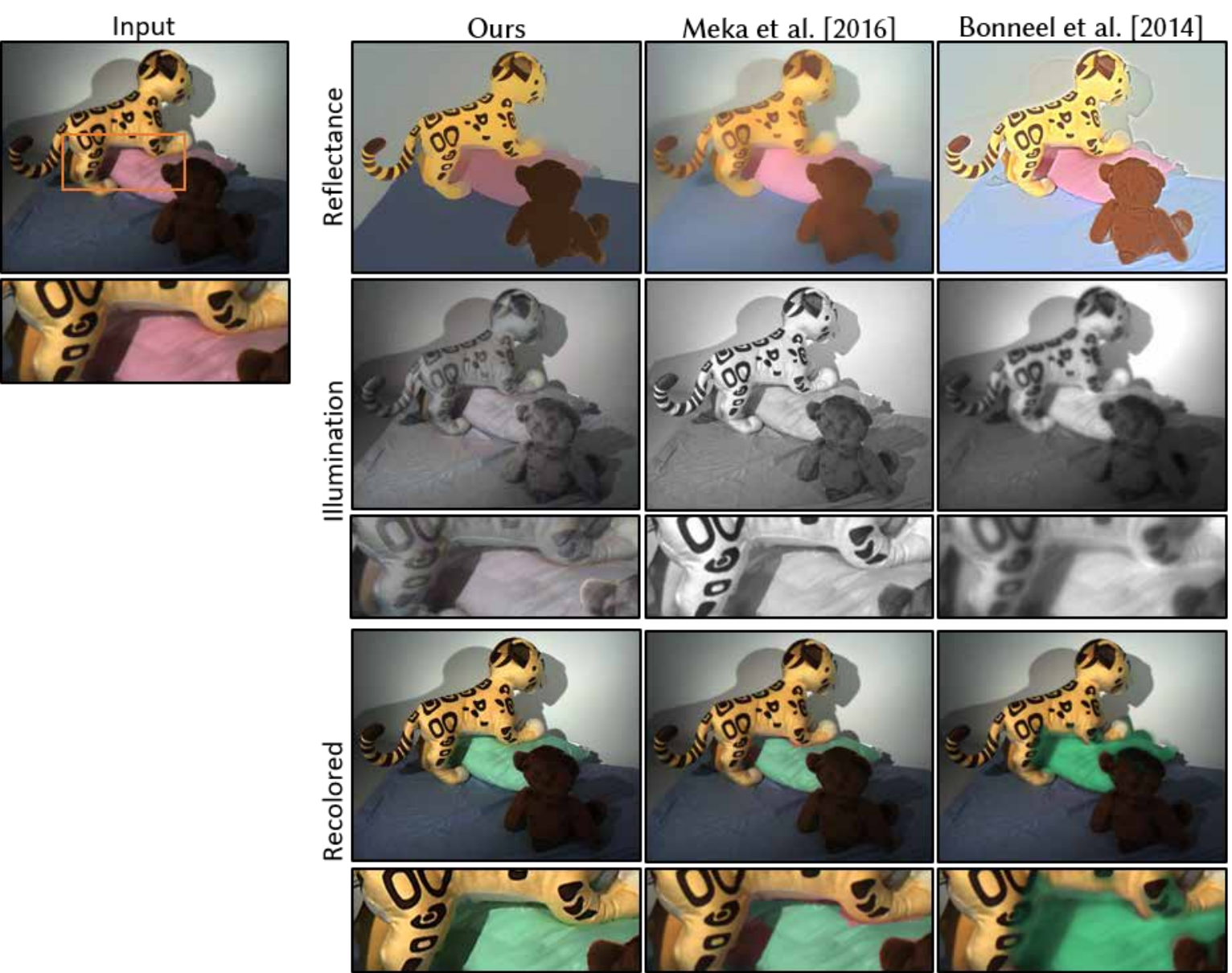}
	\caption
	{
		Comparison of our illumination decomposition to the approaches of \citet{MekaZRT2016} and \citet{BonneSTSPP2014} on the \textsc{Toys} sequence.
		With our decomposition, we achieve a higher-quality recoloring result than existing methods (see yellow arrows).
		Notice the plausible green color spill from the pillow onto the toy in our result.
	}
	\label{fig:toy}
\end{figure}

\begin{figure}
	\includegraphics[width=\linewidth]{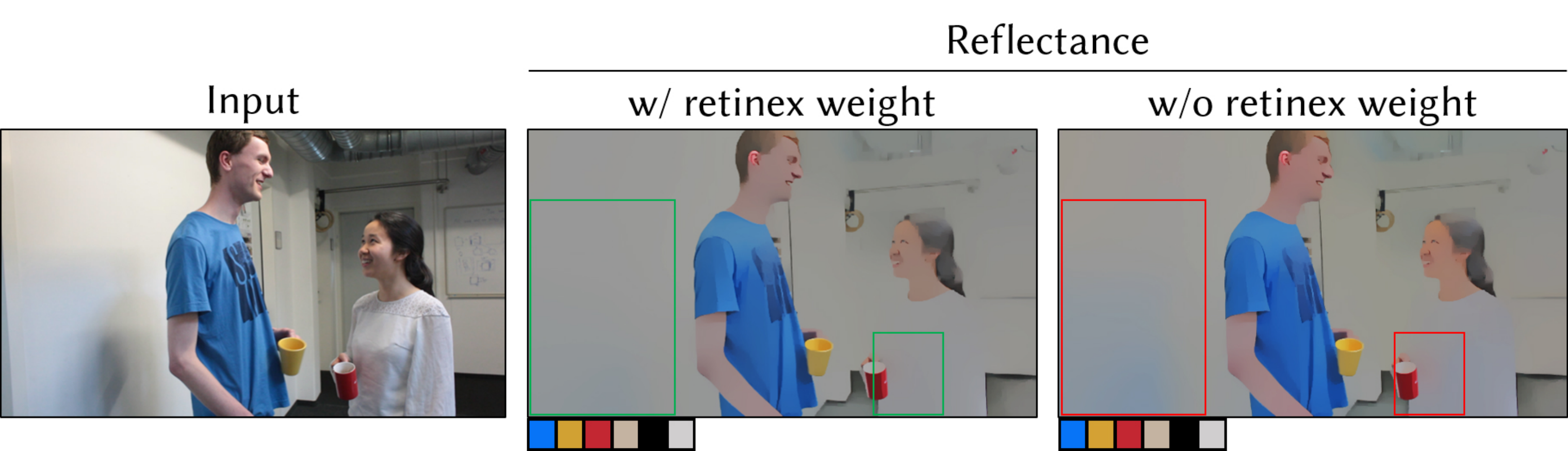}
	\caption{\label{fig:chitchat2}%
		Evaluation of the soft-color-Retinex weight of our monochromatic illumination term on the \textsc{ChitChat} sequence.
		This weight enables our approach to correctly separate the color spill on the wall and the white shirt.
	}
\end{figure}

\begin{figure*}
	\includegraphics[width=\linewidth]{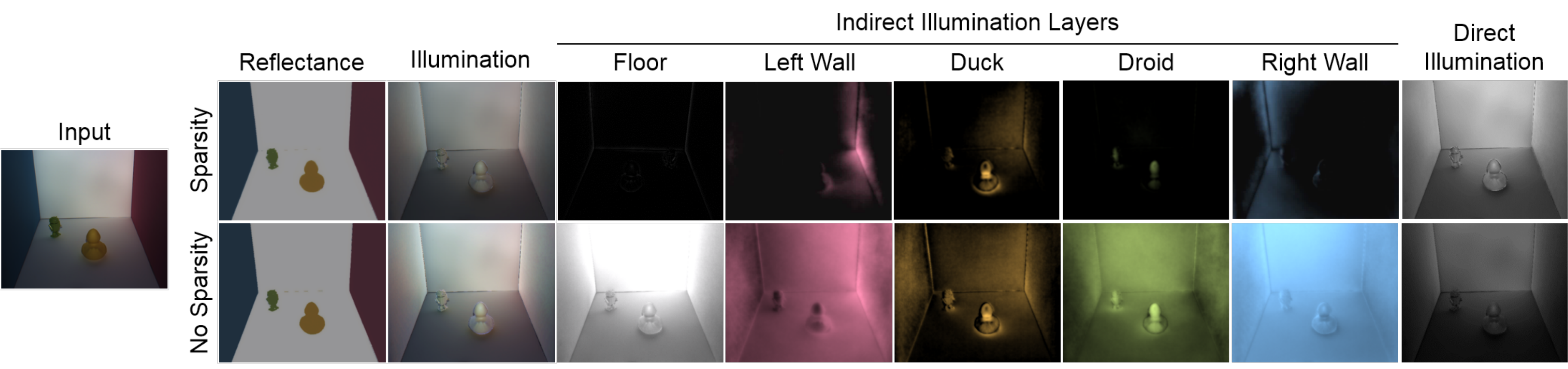}
	\caption{\label{fig:sparsity}%
		Comparison of our illumination decomposition result on the \textsc{Droid} sequence, with and without the illumination sparsity prior.
		Without the sparsity prior, the indirect illumination layers, particularly for large regions such as the walls, show activation across the entire image, which is inaccurate.
		With our sparsity prior, the contribution of the walls to the global illumination is limited to the region close to the walls and in direct sight.
	}
\end{figure*}

\begin{figure*}
	\includegraphics[width=\linewidth]{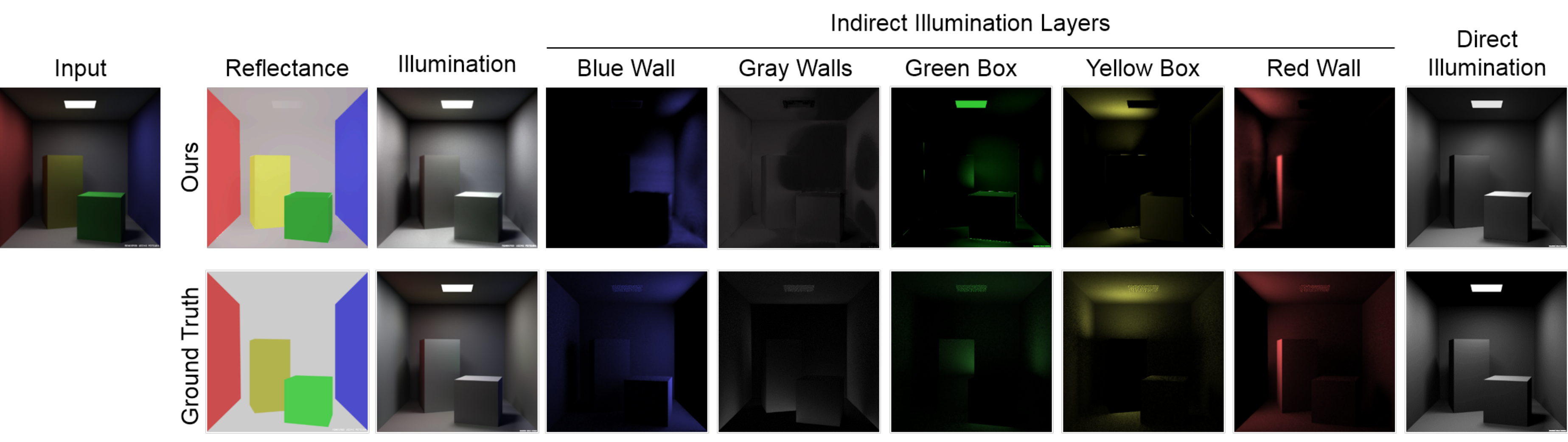}
	\caption{\label{fig:synthetic}%
		We compare our illumination decomposition qualitatively to the ground truth on the synthetic \textsc{Cornell} box.
		Our estimated indirect illumination layers capture the inter-reflections in the scene well.
		Note that we scaled the indirect illumination layers for better visualization.
	}
\end{figure*}

\begin{figure*}
	\includegraphics[width=\linewidth]{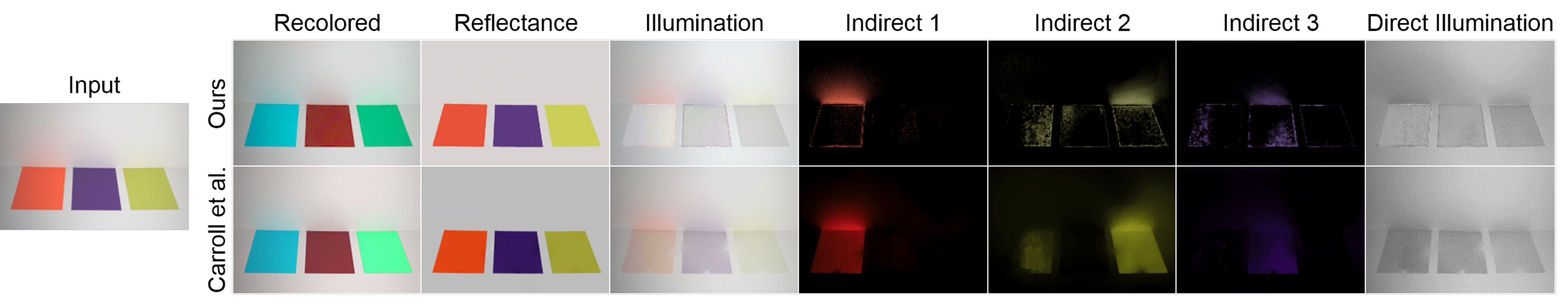}
	\caption{\label{fig:paper}%
		Comparison to \citet{CarroRA2011} on the \textsc{Paper} sequence.
		Note that their illumination image retains a lot of color in the colored paper regions, which results in a direct illumination layer that is not uniform across the table and the paper.
		Our base colors ensure that the illumination layer retains only the global illumination and not the reflectance.
		This results in sparser illumination layers, while accurately representing the color spill from the paper.
	}
\end{figure*}

\subsubsection{Evaluation of Misclustering Correction}

We evaluate our novel sparsity-based misclustering correction in \cref{fig:box,fig:toy}.
In the presence of strong inter-reflections, such as the green color spill in the shadow of the box in \cref{fig:box}, estimating the correct reflectance is highly challenging.
The state-of-the-art intrinsic decomposition approaches of \citet{MekaZRT2016} and \citet{BonneSTSPP2014} struggle in this scenario, and often miscluster the inter-reflection into the reflectance map, see \cref{fig:toy}.
This causes severe problems when an inter-reflection-consistent recoloring of the scene is required, e.g. if the green wall should be virtually replaced by a blue wall.
Our method alleviates this problem with a minimal amount of user interaction.
With a single click, the misclustered region is identified, and our approach then automatically finds the correct reflectance based on our novel correction strategy that exploits the sparsity of the indirect illumination decomposition (see \cref{sec:misclustering}).
Thus, the reflectance, direct and indirect illumination layers computed by our approach enable the seamless inter-reflection-consistent recoloring of scene elements, as shown in \cref{fig:box}.

\subsubsection{Evaluation of the Sparsity Prior}

We evaluate the importance of the sparsity prior in \cref{fig:sparsity} by
comparing our illumination decomposition result with and without the illumination sparsity prior (\cref{eq:shading-sparsity-term}).
Without the sparsity prior, the indirect illumination layers show activations across the entire image domain, which is inaccurate.
Our sparsity prior forces inter-reflections to be explained by a small number of base colors; thus the optimization has to choose how to optimally explain the inter-reflections.
This leads to sparser and more realistic indirect illumination layers that enable accurate inter-reflection-consistent recoloring.
Note that with the sparsity prior – as expected from physical light transport – the contribution of the walls to the global illumination is limited to the regions close to the walls and in direct sight.

\subsubsection{Evaluation of the Soft-Color-Retinex Weight}

We evaluate the importance of the soft-color-Retinex weight in the illumination monochromaticity prior in \cref{fig:chitchat2}.
Without the soft-Retinex weight, the prominent blue color spill on the wall and the red spill on the white shirt both incorrectly end up in the reflectance layer.
This problem is easily resolved by the soft-Retinex weight. 

\subsection{Comparisons}
\label{sec:comparisons}

We show a ground-truth comparison on synthetic data in \cref{fig:synthetic}.
In the following, we compare to the decomposition approaches of \citet{CarroRA2011}, \citet{BonneSTSPP2014} and \citet{MekaZRT2016}.

\subsubsection{Comparison to \citet{CarroRA2011}}

While our illumination layer decomposition is inspired by \citet{CarroRA2011}, it builds on top of it in a significant way, as we list below:
\begin{itemize}
  \item \citeauthor{CarroRA2011} assume that the intrinsic decomposition and base colors are given by \citet{BoussPD2009}, whereas our method jointly performs intrinsic decomposition and base color estimation with the global illumination decomposition.
  \item \citeauthor{CarroRA2011} require the user to select the base colors that contribute to the indirect illumination in the scene, whereas our clustering-based strategy automatically identifies the base colors, which are then further refined jointly with the estimation of the indirect illumination layers.
  \item \citeauthor{CarroRA2011} use the sparsity of the indirect layers to accurately perform the illumination decomposition. We additionally use this sparsity property to enhance the quality of the reflectance estimation through the base color refinement and also to automate the misclustering correction.
  \item Both methods require the user to specify the regions in which the illumination decomposition is incorrect, but their method uses strokes whereas our method uses a single click for a region, and automatically identifies the correct underlying reflectance, which is propagated across the whole region/video.
  \item  While both the methods use the IRLS strategy to solve the sparse and non-linear illumination decomposition problem, our method also optimizes for the reflectance layer and the base colors, which leads to an additional low-dimensional dense problem. We use a parallelized iterative solver to solve both problems jointly in real time. Combined with our region-tracking strategy, this enables our method to run on a live video and not just a single image.
\end{itemize}
We show a comparison in \cref{fig:paper}.
Their result retains too much color in the colored paper regions of the indirect illumination layers (\cref{fig:paper}, bottom), resulting in a direct illumination layer that is not uniform across the table and the papers.
Our base color refinement ensures that the illumination image retains only the global illumination (\cref{fig:paper}, top), and that the color variation that stems from actual surface reflectance variation is moved to the reflectance layer.
This causes our illumination layers to be more sparse, while accurately representing the color spill from the paper.
Note that we obtain these results automatically, while \citeauthor{CarroRA2011}'s approach requires several user scribbles.
Unfortunately, further comparisons on other sequences cannot be shown due to unavailability of their code/implementation for this method.

\begin{figure*}
	\includegraphics[width=\linewidth]{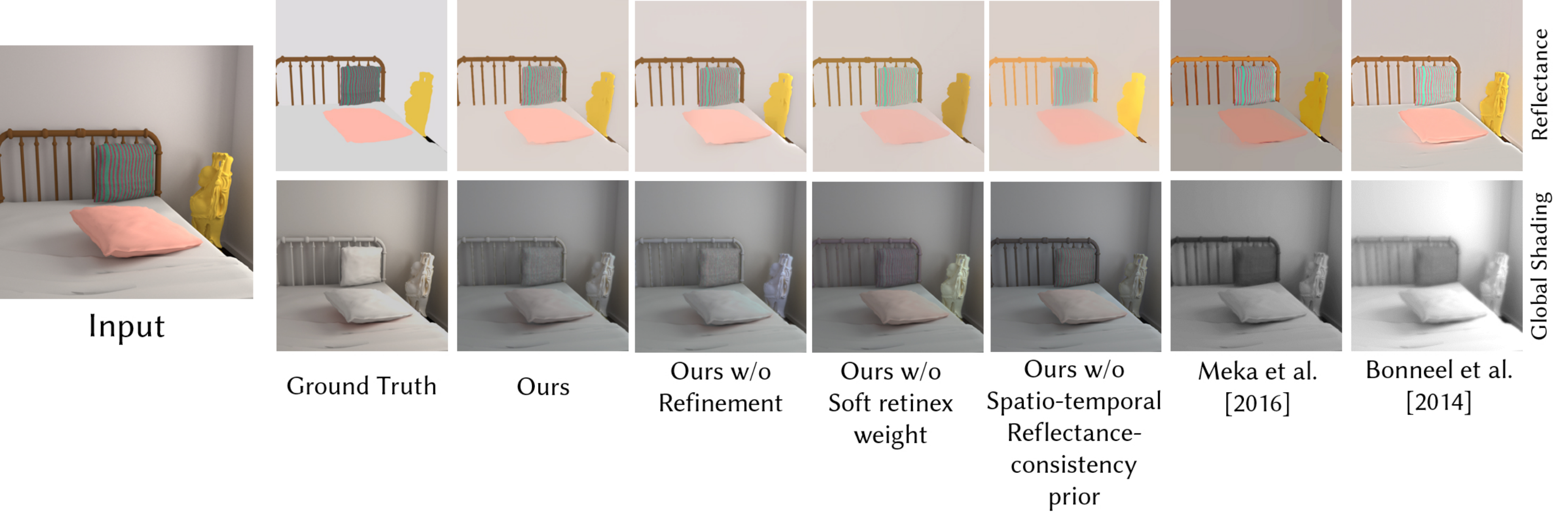}
	\caption{\label{fig:synthetic2}%
		Ablation study and qualitative comparisons on the \textsc{SyntheticRoom} sequence.
		Without the base color refinement, our approach incorrectly estimates blue illumination on the statue, the bedpost and the cushion.
		Without the soft-Retinex weighted monochromatic prior, the illumination incorrectly contains color across all the objects in the scene.
		Without the spatiotemporal prior, the reflectance estimate is inconsistent across uniform reflectance regions, such as the wall and the bedspread.
		Our result with the complete energy gives a more accurate decomposition that is closer to the ground truth.
		We accurately decompose the illumination from the pink cushion and within the yellow statue into the illumination layer, while the other intrinsic video decomposition methods incorrectly bake the color spill into the reflectance layer.
	}
\end{figure*}

\begin{figure}
	\includegraphics[width=\linewidth]{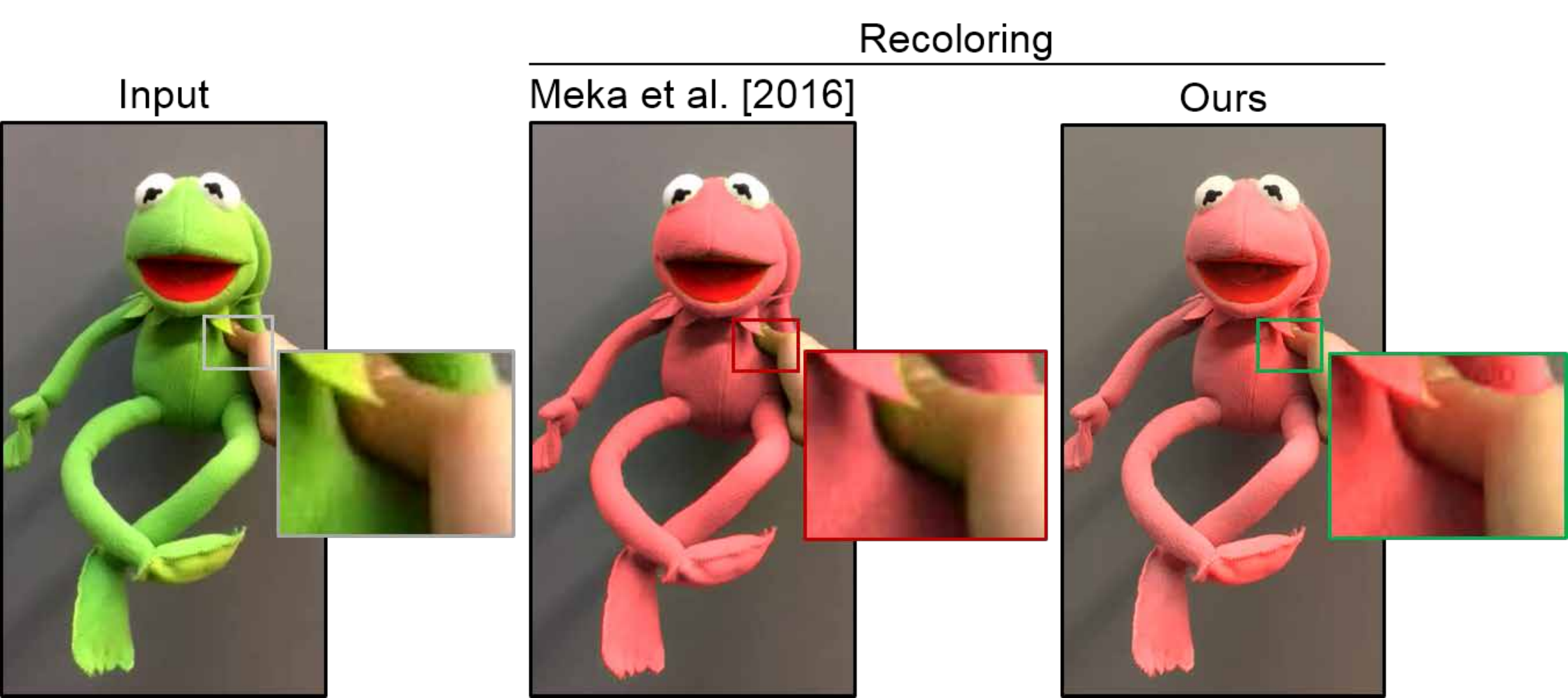}
	\caption{\label{fig:kermit1}%
		Comparison of recoloring results to \citet{MekaZRT2016} on the \textsc{Kermit} sequence \citep{BonneSTSPP2014}.
		\citeauthor{MekaZRT2016}'s approach does not correctly handle inter-reflections, e.g. from Kermit onto the thumb, while our approach consistently reconstructs and recolors these inter-reflections.
	}
\end{figure}

\subsubsection{Comparison to \citet{BonneSTSPP2014} and \citet{MekaZRT2016}}

We also compare against the intrinsic video decomposition techniques of \citet{MekaZRT2016} and \citet{BonneSTSPP2014}.
While these techniques make similar assumptions about the scene light transport as our method, they do not estimate the indirect illumination layers or the reflectance base colors.
Our joint optimization leads to more accurate decomposition as we will show in a series of comparisons.
While these two methods and our method all solve a high-dimensional non-linear optimization problem, our method also jointly solves the dense low-dimensional problem of base color refinement.
Owing to the sparsity priors on the indirect illumination layers, this gives additional priors for the reflectance estimation.

In \cref{fig:synthetic2}, we analyze our base color refinement strategy on a synthetic sequence. 
Without the refinement, the illumination is inaccurately estimated to be blueish in multiple places, which is resolved by our refinement strategy.
The other methods obtain globally inconsistent illumination results, and incorrectly bake the color spills into the reflectance layer.
In \cref{fig:kermit1}, we compare to the live intrinsic video decomposition approach of \citet{MekaZRT2016}.
Their approach does not correctly handle inter-reflections, while our approach enables inter-reflection-consistent recoloring of scene objects.
Please note the color bleeding from the green frog onto the hand.
We show a second comparison in \cref{fig:toy}, where we also compare to the off-line intrinsic video decomposition approach of \citet{BonneSTSPP2014}.
Neither of these methods is able to correctly handle scene inter-reflections.
Our supplemental videos contain additional comparisons to both methods.

\subsection{Interactive Live Applications}
\label{sec:applications}

We demonstrate several live video applications based on our illumination decomposition approach, such as inter-reflection-consistent recoloring and color keying.
For a survey of digital keying methods we refer to \citet{SchulH2006}.

\begin{figure}
	\includegraphics[width=\linewidth]{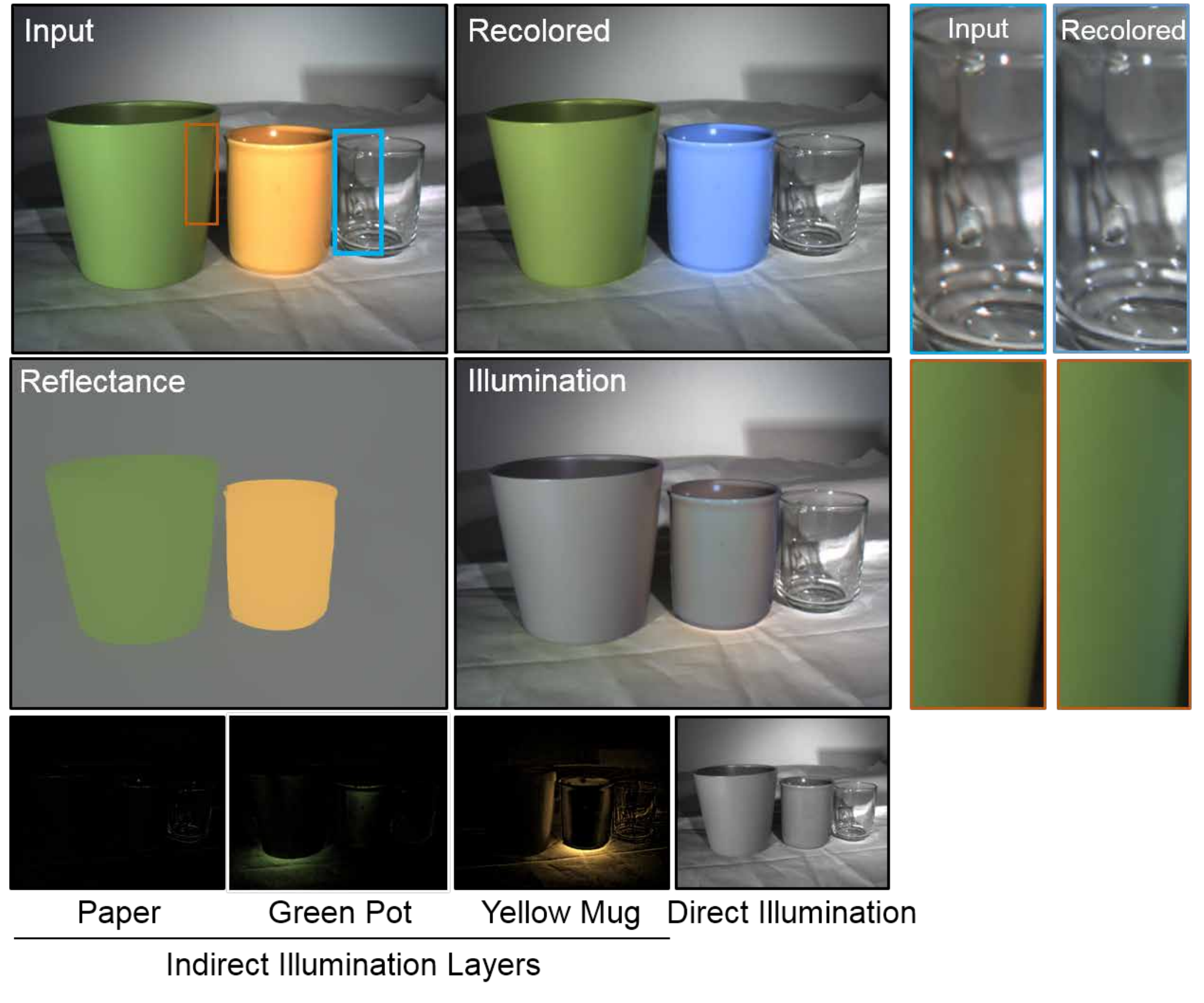}
	\caption{\label{fig:mug}%
		Recoloring result on the \textsc{Cup} sequence.
		Apart from the prominent color spills on the table cloth, even subtle inter-reflections on the green pot and the glass are captured well by our approach.
	}
\end{figure}

\subsubsection{Inter-Reflection-Consistent Recoloring}

Our illumination decomposition approach enables inter-reflection-consistent recoloring of live video streams.
We can recolor an object by modifying its associated base color, which also consistently recolors the object's  indirect illumination layer.
To recolor the reflectance map, we employ our region-growing algorithm to identify the pixels belonging to a user-selected object based on their similarity to the object's base color.
We have already shown several plausible inter-reflection-consistent recoloring results in \cref{fig:cornell2,fig:box,fig:toy,fig:kermit1}, which outperform existing intrinsic image decomposition approaches \citep{MekaZRT2016,BonneSTSPP2014}.
In \cref{fig:mug}, we further demonstrate that our approach can even recolor subtle inter-reflections on glass, and not just on diffuse surfaces.

\subsubsection{Inter-Reflection-Consistent Color Keying}

Color keying is a technique often used in visual effects for overlaying a subject in a video on top of a different background using a color-based segmentation.
In practice, a uniform green background is often used.
Global light transport in the scene often causes green inter-reflections from the background onto the subject.
This leads to unrealistic composites, since a green color spill is often visible on the subject, which does not match the new background.
Our interactive illumination approach can be used to alleviate this problem, as shown in \cref{fig:box}.
We first separate the input video into its direct and indirect illumination components.
Afterwards, we modify the base color of the green indirect illumination layer, which relights the subject to better match the new background.
This leads to more realistic outputs and can be achieved at interactive frame rates with our approach.

\begin{figure*}
	\includegraphics[width=\linewidth]{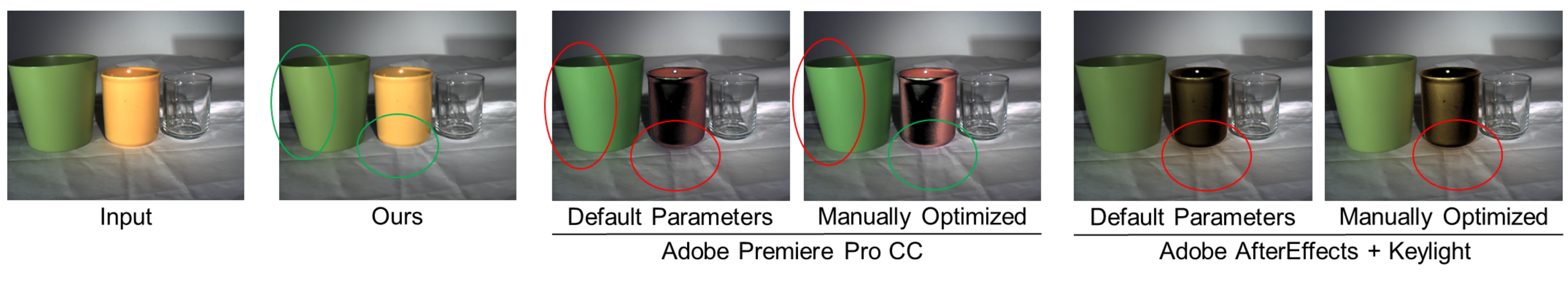}
	\caption{\label{fig:software_comparison}%
		Comparison to recoloring software.
		The commercial software cannot remove the color spill for a particular object, such as the yellow cup.
		It only supports removing a particular color component completely from the entire image.
		We show results with the default parameters and manually tuned parameters for the best results.
		Even with manually tuned parameters, the software packages cannot coherently deal with the color spill and end up introducing artifacts or inaccuracies.
		For the full sequence, we refer to the supplemental video.
	}
\end{figure*}

\subsubsection{Color-Spill Suppression}

In many video editing tasks, suppressing a strong color spill is highly important.
This technique is often used in movie and television productions to suppress the spill from a green or blue-screen.
We show an example of such an application in \cref{fig:software_comparison}.
We are able to successfully suppress the spill from the shiny yellow cup by removing the indirect illumination layer of the cup from the illumination decomposition and recombining the other layers.
We compare our results with state-of-the-art commercial software.
The tested software is not able to suppress the spill for a particular object, but only for a particular color scheme.
We also manually tuned the parameters of the software to achieve the best results.
After optimizing the parameters, Adobe Premiere Pro CC is able to suppress the spill from the cup, but it also incorrectly modifies the color of the green cup on the left side.
As is evident, our approach achieves the best results.

\section{Discussion \& Limitations}
\label{sec:discussion}

\begin{figure}
	\includegraphics[width=\linewidth]{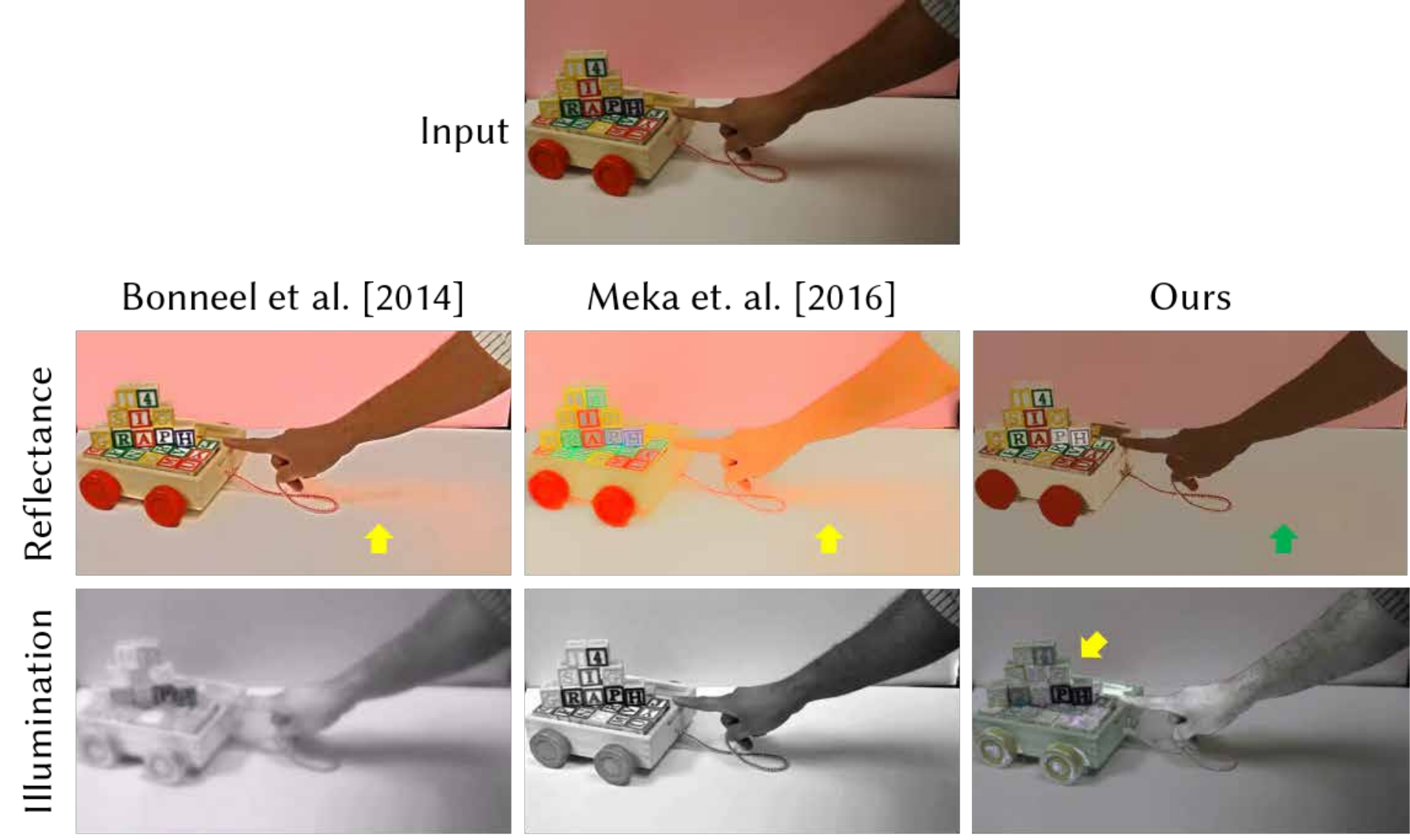}
	\caption{\label{fig:cart}%
		Comparison to state-of-the-art techniques on the \textsc{Cart} sequence.
		Note the shadow of the hand and the resulting inter-reflections on the table.
		Our technique correctly places the inter-reflections into the illumination layer, while they are baked into the reflectance layer for the other methods.
		However, due to the large number of base colors in the scene, our method incorrectly decomposes the reflectance and illumination for the blocks.
	}
\end{figure}

While we have demonstrated high-quality illumination decomposition results and a wide range of applications, our approach still has some limitations that we hope are addressed in follow-up work.
Our method operates under a few assumptions, namely, diffuse materials, smooth camera motion, white direct illumination and consistent base colors throughout the video.
Breaking these assumptions can lead to inaccurate estimations.
For example, \cref{fig:mug} shows a scene with a specular yellow mug.
Since the surface of the mug is not diffuse, our method incorrectly decomposes the specular highlight into the illumination layer.
Nonetheless, we are still able to perform a realistic recoloring of the mug.
In \cref{fig:synthetic2}, the illuminant in the scene is not white but has a yellowish hue.
While you see that in the ground truth the yellow hue is present in the illumination layer, our decomposition incorrectly places it in the reflectance layer.
This does not break down the method completely, but it can lead to slightly inaccurate results.
Solving for the color of the illuminant in the scene would require additionally solving the `color-constancy' problem, which makes the formulation even more ambiguous.
This would be an interesting avenue for future work. 

We also show in the \textsc{Box2} sequence in the supplementary video (at 03:16) that our method can face challenges if objects enter or leave the scene during the course of the video.
Modeling effects of objects that enter or leave the scene requires dynamically estimating and refining the base colors, which is computationally more challenging.

Our approach takes a monocular video as input and hence cannot model parts of the scene that are outside the view of the camera.
This means that inter-reflections caused by out-of-view objects cannot be properly modeled, since the corresponding base color might not be available.
This is a common limitation of all illumination decomposition approaches, including \citet{CarroRA2011}.

A further restriction is the user-specified upper bound on the number of base colors.
If an object with an unseen color enters the scene for the first time, and the base colors are already exceeded, its inter-reflections cannot be modeled.
This limitation could be alleviated in the future with a dynamic clustering strategy.
Quick changes in camera view or abrupt scene motion can break our region propagation strategy.
This could be alleviated by more sophisticated tracking strategies, such as SLAM.

Complex, textured scenes with many different colors are challenging to decompose, e.g. see \cref{fig:cart}, since this requires many base colors, leading to a large number of variables and an even more under-constrained optimization problem.
More sophisticated – potentially learned – scene priors could be beneficial.
Our approach only obtains plausible decompositions, since we only model light transport up to the first bounce.
Modeling the higher-order bounces would require a dramatic increase in the number of base colors, since all mixtures of reflectances would need to be considered.

More general indoor and outdoor scenes such as those in the \textit{Intrinsic Images in the Wild} dataset \citep{BellBS2014} are not the ideal use cases for our method.
This is because the scene illumination is often extremely complex, e.g., due to colored light sources and tinted windows.
Like most approaches, ours assumes white direct illumination.
Dealing with colored light sources is a more challenging problem due to the larger number of variables and thus greater ambiguity in the decomposition.
Yet, assuming some level of sparsity in the color of the light sources, the problem could still be solved using a similar formulation as ours.
We believe that this would be an interesting direction for future work.

Our decomposition results are generally temporally stable owing to the spatiotemporal reflectance consistency prior, but may exhibit some mild flickering effects at times, either due to the frequency of the indoor light sources or sudden changes in camera capture settings such as the camera aperture or auto focus or due to noise in the captured video.
Flickering in some such sequences, such as the \textsc{Box} sequence has been discussed in the supplementary video.

\section{Conclusion}
\label{sec:conclusion}

We have proposed the first illumination decomposition approach for videos.
At the core of our approach are multiple interlinked energies that enable the estimation of the direct and indirect illumination layers based on a small set of jointly estimated base colors.
The resulting decomposition problem is formulated using sparse and dense sets of non-linear equations that are solved in real time using a novel alternating data-parallel optimization strategy that is implemented on the GPU.
We have demonstrated decomposition results that qualitatively improve on existing state-of-the-art methods.
In addition, we have demonstrated various compelling appearance editing applications.
We hope that our approach will inspire follow-up work in this field.

\begin{acks}
We thank the participants of the video sequences used in this work, particularly Jiayi Wang, Lingjie Liu and Edgar Tretschk. We also thank Carroll et al. [2012] and Bonneel et al. [2014] for providing their results in their supplementary material for comparisons. We also thank the authors of `Intrinsic Images in the Wild' [Bell et al. 2014] for making their dataset available to the public. Abhimitra Meka and Christian Theobalt were supported by the ERC Consolidator Grant 4DRepLy (770784), Michael Zollhoefer was supported by Max Planck Center for Visual Computing and Communications (MPC-VCC) and Christian Richardt was supported by RCUK grant CAMERA (EP/M023281/1) and EPSRC-UKRI Innovation Fellowship (EP/S001050/1).
\end{acks}

\bibliographystyle{ACM-Reference-Format}
\bibliography{references}


\begin{thebibliography}{65}


\ifx \showCODEN    \undefined \def \showCODEN     #1{\unskip}     \fi
\ifx \showDOI      \undefined \def \showDOI       #1{#1}\fi
\ifx \showISBNx    \undefined \def \showISBNx     #1{\unskip}     \fi
\ifx \showISBNxiii \undefined \def \showISBNxiii  #1{\unskip}     \fi
\ifx \showISSN     \undefined \def \showISSN      #1{\unskip}     \fi
\ifx \showLCCN     \undefined \def \showLCCN      #1{\unskip}     \fi
\ifx \shownote     \undefined \def \shownote      #1{#1}          \fi
\ifx \showarticletitle \undefined \def \showarticletitle #1{#1}   \fi
\ifx \showURL      \undefined \def \showURL       {\relax}        \fi
\providecommand\bibfield[2]{#2}
\providecommand\bibinfo[2]{#2}
\providecommand\natexlab[1]{#1}
\providecommand\showeprint[2][]{arXiv:#2}

\bibitem[\protect\citeauthoryear{Aksoy, Aydın, Pollefeys, and Smolić}{Aksoy
  et~al\mbox{.}}{2016}]%
        {AksoyAPS2016}
\bibfield{author}{\bibinfo{person}{Yağız Aksoy}, \bibinfo{person}{Tunç~Ozan
  Aydın}, \bibinfo{person}{Marc Pollefeys}, {and} \bibinfo{person}{Aljoša
  Smolić}.} \bibinfo{year}{2016}\natexlab{}.
\newblock \showarticletitle{Interactive high-quality {green-screen} keying via
  color unmixing}.
\newblock \bibinfo{journal}{\emph{ACM Trans. Graph.}} \bibinfo{volume}{35},
  \bibinfo{number}{5} (\bibinfo{date}{August} \bibinfo{year}{2016}),
  \bibinfo{pages}{152:1--12}.
\newblock
\urldef\tempurl%
\url{https://doi.org/10.1145/2907940}
\showDOI{\tempurl}


\bibitem[\protect\citeauthoryear{Aksoy, Aydın, Smolić, and Pollefeys}{Aksoy
  et~al\mbox{.}}{2017}]%
        {AksoyASP2017}
\bibfield{author}{\bibinfo{person}{Yağız Aksoy}, \bibinfo{person}{Tunç~Ozan
  Aydın}, \bibinfo{person}{Aljoša Smolić}, {and} \bibinfo{person}{Marc
  Pollefeys}.} \bibinfo{year}{2017}\natexlab{}.
\newblock \showarticletitle{Unmixing-Based Soft Color Segmentation for Image
  Manipulation}.
\newblock \bibinfo{journal}{\emph{ACM Trans. Graph.}} \bibinfo{volume}{36},
  \bibinfo{number}{2} (\bibinfo{date}{March} \bibinfo{year}{2017}),
  \bibinfo{pages}{19:1--19}.
\newblock
\urldef\tempurl%
\url{https://doi.org/10.1145/3002176}
\showDOI{\tempurl}


\bibitem[\protect\citeauthoryear{Alperovich and Goldluecke}{Alperovich and
  Goldluecke}{2017}]%
        {AlperG2017}
\bibfield{author}{\bibinfo{person}{Anna Alperovich} {and}
  \bibinfo{person}{Bastian Goldluecke}.} \bibinfo{year}{2017}\natexlab{}.
\newblock \showarticletitle{A Variational Model for Intrinsic Light Field
  Decomposition}. In \bibinfo{booktitle}{\emph{ACCV}}. \bibinfo{pages}{66--82}.
\newblock
\urldef\tempurl%
\url{https://doi.org/10.1007/978-3-319-54187-7_5}
\showDOI{\tempurl}


\bibitem[\protect\citeauthoryear{Azinović, Li, Kaplanyan, and
  Nießner}{Azinović et~al\mbox{.}}{2019}]%
        {AzinoLKN2019}
\bibfield{author}{\bibinfo{person}{Dejan Azinović}, \bibinfo{person}{Tzu-Mao
  Li}, \bibinfo{person}{Anton Kaplanyan}, {and} \bibinfo{person}{Matthias
  Nießner}.} \bibinfo{year}{2019}\natexlab{}.
\newblock \showarticletitle{Inverse Path Tracing for Joint Material and
  Lighting Estimation}. In \bibinfo{booktitle}{\emph{CVPR}}.
  \bibinfo{pages}{2442--2451}.
\newblock
\urldef\tempurl%
\url{https://doi.org/10.1109/CVPR.2019.00255}
\showDOI{\tempurl}


\bibitem[\protect\citeauthoryear{Bach, Jenatton, Mairal, and Obozinski}{Bach
  et~al\mbox{.}}{2012}]%
        {BachJMO2012}
\bibfield{author}{\bibinfo{person}{Francis Bach}, \bibinfo{person}{Rodolphe
  Jenatton}, \bibinfo{person}{Julien Mairal}, {and} \bibinfo{person}{Guillaume
  Obozinski}.} \bibinfo{year}{2012}\natexlab{}.
\newblock \showarticletitle{Optimization with Sparsity-Inducing Penalties}.
\newblock \bibinfo{journal}{\emph{Foundations and Trends in Machine Learning}}
  \bibinfo{volume}{4}, \bibinfo{number}{1} (\bibinfo{year}{2012}),
  \bibinfo{pages}{1--106}.
\newblock
\urldef\tempurl%
\url{https://doi.org/10.1561/2200000015}
\showDOI{\tempurl}


\bibitem[\protect\citeauthoryear{Barron and Malik}{Barron and Malik}{2015}]%
        {BarroM2015}
\bibfield{author}{\bibinfo{person}{Jonathan~T. Barron} {and}
  \bibinfo{person}{Jitendra Malik}.} \bibinfo{year}{2015}\natexlab{}.
\newblock \showarticletitle{Shape, Illumination, and Reflectance from Shading}.
\newblock \bibinfo{journal}{\emph{IEEE Trans. Pattern Anal. Mach. Intell.}}
  \bibinfo{volume}{37}, \bibinfo{number}{8} (\bibinfo{date}{August}
  \bibinfo{year}{2015}), \bibinfo{pages}{1670--1687}.
\newblock
\urldef\tempurl%
\url{https://doi.org/10.1109/TPAMI.2014.2377712}
\showDOI{\tempurl}


\bibitem[\protect\citeauthoryear{Barrow and Tenenbaum}{Barrow and
  Tenenbaum}{1978}]%
        {BarroT1978}
\bibfield{author}{\bibinfo{person}{Harry~G. Barrow} {and}
  \bibinfo{person}{Jay~M. Tenenbaum}.} \bibinfo{year}{1978}\natexlab{}.
\newblock \showarticletitle{Recovering intrinsic scene characteristics from
  images}. In \bibinfo{booktitle}{\emph{Computer Vision Systems}}.
  \bibinfo{pages}{3--26}.
\newblock


\bibitem[\protect\citeauthoryear{Beigpour and van~de Weijer}{Beigpour and
  van~de Weijer}{2011}]%
        {BeigpW2011}
\bibfield{author}{\bibinfo{person}{Shida Beigpour} {and} \bibinfo{person}{Joost
  van~de Weijer}.} \bibinfo{year}{2011}\natexlab{}.
\newblock \showarticletitle{Object recoloring based on intrinsic image
  estimation}. In \bibinfo{booktitle}{\emph{ICCV}}. \bibinfo{pages}{327--334}.
\newblock
\urldef\tempurl%
\url{https://doi.org/10.1109/ICCV.2011.6126259}
\showDOI{\tempurl}


\bibitem[\protect\citeauthoryear{Bell, Bala, and Snavely}{Bell
  et~al\mbox{.}}{2014}]%
        {BellBS2014}
\bibfield{author}{\bibinfo{person}{Sean Bell}, \bibinfo{person}{Kavita Bala},
  {and} \bibinfo{person}{Noah Snavely}.} \bibinfo{year}{2014}\natexlab{}.
\newblock \showarticletitle{Intrinsic Images in the Wild}.
\newblock \bibinfo{journal}{\emph{ACM Trans. Graph.}} \bibinfo{volume}{33},
  \bibinfo{number}{4} (\bibinfo{date}{July} \bibinfo{year}{2014}),
  \bibinfo{pages}{159:1--12}.
\newblock
\urldef\tempurl%
\url{https://doi.org/10.1145/2601097.2601206}
\showDOI{\tempurl}


\bibitem[\protect\citeauthoryear{Bi, Han, and Yu}{Bi et~al\mbox{.}}{2015}]%
        {BiHY2015}
\bibfield{author}{\bibinfo{person}{Sai Bi}, \bibinfo{person}{Xiaoguang Han},
  {and} \bibinfo{person}{Yizhou Yu}.} \bibinfo{year}{2015}\natexlab{}.
\newblock \showarticletitle{An $L_1$ Image Transform for Edge-Preserving
  Smoothing and Scene-Level Intrinsic Decomposition}.
\newblock \bibinfo{journal}{\emph{ACM Trans. Graph.}} \bibinfo{volume}{34},
  \bibinfo{number}{4} (\bibinfo{date}{July} \bibinfo{year}{2015}),
  \bibinfo{pages}{78:1--12}.
\newblock
\urldef\tempurl%
\url{https://doi.org/10.1145/2766946}
\showDOI{\tempurl}


\bibitem[\protect\citeauthoryear{Bi, Kalantari, and Ramamoorthi}{Bi
  et~al\mbox{.}}{2018}]%
        {BiKR2018}
\bibfield{author}{\bibinfo{person}{Sai Bi}, \bibinfo{person}{Nima~Khademi
  Kalantari}, {and} \bibinfo{person}{Ravi Ramamoorthi}.}
  \bibinfo{year}{2018}\natexlab{}.
\newblock \showarticletitle{Deep Hybrid Real and Synthetic Training for
  Intrinsic Decomposition}. In \bibinfo{booktitle}{\emph{Eurographics Symposium
  on Rendering -- Experimental Ideas \& Implementations}}.
\newblock


\bibitem[\protect\citeauthoryear{Bonneel, Kovacs, Paris, and Bala}{Bonneel
  et~al\mbox{.}}{2017}]%
        {BonneKPB2017}
\bibfield{author}{\bibinfo{person}{Nicolas Bonneel}, \bibinfo{person}{Balazs
  Kovacs}, \bibinfo{person}{Sylvain Paris}, {and} \bibinfo{person}{Kavita
  Bala}.} \bibinfo{year}{2017}\natexlab{}.
\newblock \showarticletitle{Intrinsic Decompositions for Image Editing}.
\newblock \bibinfo{journal}{\emph{Comput. Graph. Forum}} \bibinfo{volume}{36},
  \bibinfo{number}{2} (\bibinfo{date}{May} \bibinfo{year}{2017}),
  \bibinfo{pages}{593--609}.
\newblock
\urldef\tempurl%
\url{https://doi.org/10.1111/cgf.13149}
\showDOI{\tempurl}


\bibitem[\protect\citeauthoryear{Bonneel, Sunkavalli, Tompkin, Sun, Paris, and
  Pfister}{Bonneel et~al\mbox{.}}{2014}]%
        {BonneSTSPP2014}
\bibfield{author}{\bibinfo{person}{Nicolas Bonneel}, \bibinfo{person}{Kalyan
  Sunkavalli}, \bibinfo{person}{James Tompkin}, \bibinfo{person}{Deqing Sun},
  \bibinfo{person}{Sylvain Paris}, {and} \bibinfo{person}{Hanspeter Pfister}.}
  \bibinfo{year}{2014}\natexlab{}.
\newblock \showarticletitle{Interactive Intrinsic Video Editing}.
\newblock \bibinfo{journal}{\emph{ACM Trans. Graph.}} \bibinfo{volume}{33},
  \bibinfo{number}{6} (\bibinfo{date}{November} \bibinfo{year}{2014}),
  \bibinfo{pages}{197:1--10}.
\newblock
\urldef\tempurl%
\url{https://doi.org/10.1145/2661229.2661253}
\showDOI{\tempurl}


\bibitem[\protect\citeauthoryear{Bonneel, Tompkin, Sunkavalli, Sun, Paris, and
  Pfister}{Bonneel et~al\mbox{.}}{2015}]%
        {BonneTSSPP2015}
\bibfield{author}{\bibinfo{person}{Nicolas Bonneel}, \bibinfo{person}{James
  Tompkin}, \bibinfo{person}{Kalyan Sunkavalli}, \bibinfo{person}{Deqing Sun},
  \bibinfo{person}{Sylvain Paris}, {and} \bibinfo{person}{Hanspeter Pfister}.}
  \bibinfo{year}{2015}\natexlab{}.
\newblock \showarticletitle{Blind Video Temporal Consistency}.
\newblock \bibinfo{journal}{\emph{ACM Trans. Graph.}} \bibinfo{volume}{34},
  \bibinfo{number}{6} (\bibinfo{date}{November} \bibinfo{year}{2015}),
  \bibinfo{pages}{196:1--9}.
\newblock
\urldef\tempurl%
\url{https://doi.org/10.1145/2816795.2818107}
\showDOI{\tempurl}


\bibitem[\protect\citeauthoryear{Bousseau, Paris, and Durand}{Bousseau
  et~al\mbox{.}}{2009}]%
        {BoussPD2009}
\bibfield{author}{\bibinfo{person}{Adrien Bousseau}, \bibinfo{person}{Sylvain
  Paris}, {and} \bibinfo{person}{Frédo Durand}.}
  \bibinfo{year}{2009}\natexlab{}.
\newblock \showarticletitle{User-Assisted Intrinsic Images}.
\newblock \bibinfo{journal}{\emph{ACM Trans. Graph.}} \bibinfo{volume}{28},
  \bibinfo{number}{5} (\bibinfo{date}{December} \bibinfo{year}{2009}),
  \bibinfo{pages}{130:1--10}.
\newblock
\urldef\tempurl%
\url{https://doi.org/10.1145/1618452.1618476}
\showDOI{\tempurl}


\bibitem[\protect\citeauthoryear{Carroll, Ramamoorthi, and Agrawala}{Carroll
  et~al\mbox{.}}{2011}]%
        {CarroRA2011}
\bibfield{author}{\bibinfo{person}{Robert Carroll}, \bibinfo{person}{Ravi
  Ramamoorthi}, {and} \bibinfo{person}{Maneesh Agrawala}.}
  \bibinfo{year}{2011}\natexlab{}.
\newblock \showarticletitle{Illumination decomposition for material recoloring
  with consistent interreflections}.
\newblock \bibinfo{journal}{\emph{ACM Trans. Graph.}} \bibinfo{volume}{30},
  \bibinfo{number}{4} (\bibinfo{date}{July} \bibinfo{year}{2011}),
  \bibinfo{pages}{43:1--10}.
\newblock
\urldef\tempurl%
\url{https://doi.org/10.1145/2010324.1964938}
\showDOI{\tempurl}


\bibitem[\protect\citeauthoryear{Chang, Cabezas, and Fisher}{Chang
  et~al\mbox{.}}{2014}]%
        {ChangCF2014}
\bibfield{author}{\bibinfo{person}{Jason Chang}, \bibinfo{person}{Randi
  Cabezas}, {and} \bibinfo{person}{John~W. Fisher, III}.}
  \bibinfo{year}{2014}\natexlab{}.
\newblock \showarticletitle{{Bayesian} Nonparametric Intrinsic Image
  Decomposition}. In \bibinfo{booktitle}{\emph{ECCV}},
  Vol.~\bibinfo{volume}{8692}. \bibinfo{pages}{704--719}.
\newblock
\urldef\tempurl%
\url{https://doi.org/10.1007/978-3-319-10593-2_46}
\showDOI{\tempurl}


\bibitem[\protect\citeauthoryear{DeVito, Mara, Zollhöfer, Bernstein,
  Ragan-Kelley, Theobalt, Hanrahan, Fisher, and Nießner}{DeVito
  et~al\mbox{.}}{2017}]%
        {DeVitMZBRTHFN2017}
\bibfield{author}{\bibinfo{person}{Zachary DeVito}, \bibinfo{person}{Michael
  Mara}, \bibinfo{person}{Michael Zollhöfer}, \bibinfo{person}{Gilbert
  Bernstein}, \bibinfo{person}{Jonathan Ragan-Kelley},
  \bibinfo{person}{Christian Theobalt}, \bibinfo{person}{Pat Hanrahan},
  \bibinfo{person}{Matthew Fisher}, {and} \bibinfo{person}{Matthias Nießner}.}
  \bibinfo{year}{2017}\natexlab{}.
\newblock \showarticletitle{Opt: A Domain Specific Language for Non-Linear
  Least Squares Optimization in Graphics and Imaging}.
\newblock \bibinfo{journal}{\emph{ACM Trans. Graph.}} \bibinfo{volume}{36},
  \bibinfo{number}{5} (\bibinfo{date}{October} \bibinfo{year}{2017}),
  \bibinfo{pages}{171:1--27}.
\newblock
\urldef\tempurl%
\url{https://doi.org/10.1145/3132188}
\showDOI{\tempurl}


\bibitem[\protect\citeauthoryear{Ding, Sheng, Hou, Xie, and Ma}{Ding
  et~al\mbox{.}}{2017}]%
        {DingSHXM2017}
\bibfield{author}{\bibinfo{person}{Shouhong Ding}, \bibinfo{person}{Bin Sheng},
  \bibinfo{person}{Xiaonan Hou}, \bibinfo{person}{Zhifeng Xie}, {and}
  \bibinfo{person}{Lizhuang Ma}.} \bibinfo{year}{2017}\natexlab{}.
\newblock \showarticletitle{Intrinsic Image Decomposition Using Multi-Scale
  Measurements and Sparsity}.
\newblock \bibinfo{journal}{\emph{Comput. Graph. Forum}} \bibinfo{volume}{36},
  \bibinfo{number}{6} (\bibinfo{year}{2017}), \bibinfo{pages}{251--261}.
\newblock
\urldef\tempurl%
\url{https://doi.org/10.1111/cgf.12874}
\showDOI{\tempurl}


\bibitem[\protect\citeauthoryear{Dong, Dong, Tong, and Peers}{Dong
  et~al\mbox{.}}{2015}]%
        {DongDTP2015}
\bibfield{author}{\bibinfo{person}{Bo Dong}, \bibinfo{person}{Yue Dong},
  \bibinfo{person}{Xin Tong}, {and} \bibinfo{person}{Pieter Peers}.}
  \bibinfo{year}{2015}\natexlab{}.
\newblock \showarticletitle{Measurement-based Editing of Diffuse Albedo with
  Consistent Interreflections}.
\newblock \bibinfo{journal}{\emph{ACM Trans. Graph.}} \bibinfo{volume}{34},
  \bibinfo{number}{4} (\bibinfo{date}{July} \bibinfo{year}{2015}),
  \bibinfo{pages}{112:1--11}.
\newblock
\urldef\tempurl%
\url{https://doi.org/10.1145/2766979}
\showDOI{\tempurl}


\bibitem[\protect\citeauthoryear{Dong, Chen, Peers, Zhang, and Tong}{Dong
  et~al\mbox{.}}{2014}]%
        {DongCPZT2014}
\bibfield{author}{\bibinfo{person}{Yue Dong}, \bibinfo{person}{Guojun Chen},
  \bibinfo{person}{Pieter Peers}, \bibinfo{person}{Jiawan Zhang}, {and}
  \bibinfo{person}{Xin Tong}.} \bibinfo{year}{2014}\natexlab{}.
\newblock \showarticletitle{Appearance-from-motion: Recovering Spatially
  Varying Surface Reflectance Under Unknown Lighting}.
\newblock \bibinfo{journal}{\emph{ACM Trans. Graph.}} \bibinfo{volume}{33},
  \bibinfo{number}{6} (\bibinfo{date}{November} \bibinfo{year}{2014}),
  \bibinfo{pages}{193:1--12}.
\newblock
\urldef\tempurl%
\url{https://doi.org/10.1145/2661229.2661283}
\showDOI{\tempurl}


\bibitem[\protect\citeauthoryear{Favreau, Lafarge, and Bousseau}{Favreau
  et~al\mbox{.}}{2017}]%
        {FavreLB2017}
\bibfield{author}{\bibinfo{person}{Jean-Dominique Favreau},
  \bibinfo{person}{Florent Lafarge}, {and} \bibinfo{person}{Adrien Bousseau}.}
  \bibinfo{year}{2017}\natexlab{}.
\newblock \showarticletitle{{Photo2ClipArt}: Image Abstraction and
  Vectorization Using Layered Linear Gradients}.
\newblock \bibinfo{journal}{\emph{ACM Trans. Graph.}} \bibinfo{volume}{36},
  \bibinfo{number}{6} (\bibinfo{date}{November} \bibinfo{year}{2017}),
  \bibinfo{pages}{180:1--11}.
\newblock
\urldef\tempurl%
\url{https://doi.org/10.1145/3130800.3130888}
\showDOI{\tempurl}


\bibitem[\protect\citeauthoryear{Garces, Echevarria, Zhang, Wu, Zhou, and
  Gutierrez}{Garces et~al\mbox{.}}{2017}]%
        {GarceEZWZG2017}
\bibfield{author}{\bibinfo{person}{Elena Garces}, \bibinfo{person}{Jose~I.
  Echevarria}, \bibinfo{person}{Wen Zhang}, \bibinfo{person}{Hongzhi Wu},
  \bibinfo{person}{Kun Zhou}, {and} \bibinfo{person}{Diego Gutierrez}.}
  \bibinfo{year}{2017}\natexlab{}.
\newblock \showarticletitle{Intrinsic Light Field Images}.
\newblock \bibinfo{journal}{\emph{Comput. Graph. Forum}} \bibinfo{volume}{36},
  \bibinfo{number}{8} (\bibinfo{date}{December} \bibinfo{year}{2017}),
  \bibinfo{pages}{589--599}.
\newblock
\urldef\tempurl%
\url{https://doi.org/10.1111/cgf.13154}
\showDOI{\tempurl}


\bibitem[\protect\citeauthoryear{Garces, Muñoz, Lopez-Moreno, and
  Gutierrez}{Garces et~al\mbox{.}}{2012}]%
        {GarceMLG2012}
\bibfield{author}{\bibinfo{person}{Elena Garces}, \bibinfo{person}{Adolfo
  Muñoz}, \bibinfo{person}{Jorge Lopez-Moreno}, {and} \bibinfo{person}{Diego
  Gutierrez}.} \bibinfo{year}{2012}\natexlab{}.
\newblock \showarticletitle{Intrinsic Images by Clustering}.
\newblock \bibinfo{journal}{\emph{Comput. Graph. Forum}} \bibinfo{volume}{31},
  \bibinfo{number}{4} (\bibinfo{year}{2012}), \bibinfo{pages}{1415--1424}.
\newblock
\urldef\tempurl%
\url{https://doi.org/10.1111/j.1467-8659.2012.03137.x}
\showDOI{\tempurl}


\bibitem[\protect\citeauthoryear{Georgoulis, Rematas, Ritschel, Gavves, Fritz,
  Van~Gool, and Tuytelaars}{Georgoulis et~al\mbox{.}}{2018}]%
        {GeorgRRGFVT2018}
\bibfield{author}{\bibinfo{person}{Stamatios Georgoulis},
  \bibinfo{person}{Konstantinos Rematas}, \bibinfo{person}{Tobias Ritschel},
  \bibinfo{person}{Efstratios Gavves}, \bibinfo{person}{Mario Fritz},
  \bibinfo{person}{Luc Van~Gool}, {and} \bibinfo{person}{Tinne Tuytelaars}.}
  \bibinfo{year}{2018}\natexlab{}.
\newblock \showarticletitle{Reflectance and Natural Illumination from
  Single-Material Specular Objects Using Deep Learning}.
\newblock \bibinfo{journal}{\emph{IEEE Trans. Pattern Anal. Mach. Intell.}}
  \bibinfo{volume}{40}, \bibinfo{number}{8} (\bibinfo{date}{August}
  \bibinfo{year}{2018}), \bibinfo{pages}{1932--1947}.
\newblock
\urldef\tempurl%
\url{https://doi.org/10.1109/TPAMI.2017.2742999}
\showDOI{\tempurl}


\bibitem[\protect\citeauthoryear{Grosse, Johnson, Adelson, and Freeman}{Grosse
  et~al\mbox{.}}{2009}]%
        {GrossJAF2009}
\bibfield{author}{\bibinfo{person}{Roger Grosse}, \bibinfo{person}{Micah~K.
  Johnson}, \bibinfo{person}{Edward~H. Adelson}, {and}
  \bibinfo{person}{William~T. Freeman}.} \bibinfo{year}{2009}\natexlab{}.
\newblock \showarticletitle{Ground truth dataset and baseline evaluations for
  intrinsic image algorithms}. In \bibinfo{booktitle}{\emph{ICCV}}.
  \bibinfo{pages}{2335--2342}.
\newblock
\urldef\tempurl%
\url{https://doi.org/10.1109/ICCV.2009.5459428}
\showDOI{\tempurl}


\bibitem[\protect\citeauthoryear{Guo, Xu, Yu, Liu, Dai, and Liu}{Guo
  et~al\mbox{.}}{2017}]%
        {GuoXYLDL2017}
\bibfield{author}{\bibinfo{person}{Kaiwen Guo}, \bibinfo{person}{Feng Xu},
  \bibinfo{person}{Tao Yu}, \bibinfo{person}{Xiaoyang Liu},
  \bibinfo{person}{Qionghai Dai}, {and} \bibinfo{person}{Yebin Liu}.}
  \bibinfo{year}{2017}\natexlab{}.
\newblock \showarticletitle{Real-time Geometry, Albedo and Motion
  Reconstruction Using a Single {RGBD} Camera}.
\newblock \bibinfo{journal}{\emph{ACM Trans. Graph.}} \bibinfo{volume}{36},
  \bibinfo{number}{3} (\bibinfo{date}{June} \bibinfo{year}{2017}),
  \bibinfo{pages}{32:1--13}.
\newblock
\urldef\tempurl%
\url{https://doi.org/10.1145/3083722}
\showDOI{\tempurl}


\bibitem[\protect\citeauthoryear{Holland and Welsch}{Holland and
  Welsch}{1977}]%
        {HollaW1977}
\bibfield{author}{\bibinfo{person}{Paul~W. Holland} {and}
  \bibinfo{person}{Roy~E. Welsch}.} \bibinfo{year}{1977}\natexlab{}.
\newblock \showarticletitle{Robust regression using iteratively reweighted
  least-squares}.
\newblock \bibinfo{journal}{\emph{Communications in Statistics – Theory and
  Methods}} \bibinfo{volume}{6}, \bibinfo{number}{9} (\bibinfo{date}{September}
  \bibinfo{year}{1977}), \bibinfo{pages}{813--827}.
\newblock
\urldef\tempurl%
\url{https://doi.org/10.1080/03610927708827533}
\showDOI{\tempurl}


\bibitem[\protect\citeauthoryear{Innamorati, Ritschel, Weyrich, and
  Mitra}{Innamorati et~al\mbox{.}}{2017}]%
        {InnamRWM2017}
\bibfield{author}{\bibinfo{person}{Carlo Innamorati}, \bibinfo{person}{Tobias
  Ritschel}, \bibinfo{person}{Tim Weyrich}, {and} \bibinfo{person}{Niloy~J.
  Mitra}.} \bibinfo{year}{2017}\natexlab{}.
\newblock \showarticletitle{Decomposing Single Images for Layered Photo
  Retouching}.
\newblock \bibinfo{journal}{\emph{Comput. Graph. Forum}} \bibinfo{volume}{36},
  \bibinfo{number}{4} (\bibinfo{date}{July} \bibinfo{year}{2017}),
  \bibinfo{pages}{15--25}.
\newblock
\urldef\tempurl%
\url{https://doi.org/10.1111/cgf.13220}
\showDOI{\tempurl}


\bibitem[\protect\citeauthoryear{Janner, Wu, Kulkarni, Yildirim, and
  Tenenbaum}{Janner et~al\mbox{.}}{2017}]%
        {JanneWKYT2017}
\bibfield{author}{\bibinfo{person}{Michael Janner}, \bibinfo{person}{Jiajun
  Wu}, \bibinfo{person}{Tejas~D. Kulkarni}, \bibinfo{person}{Ilker Yildirim},
  {and} \bibinfo{person}{Joshua~B. Tenenbaum}.}
  \bibinfo{year}{2017}\natexlab{}.
\newblock \showarticletitle{Self-Supervised Intrinsic Image Decomposition}. In
  \bibinfo{booktitle}{\emph{NIPS}}.
\newblock
\urldef\tempurl%
\url{http://rin.csail.mit.edu/}
\showURL{%
\tempurl}


\bibitem[\protect\citeauthoryear{Kajiya}{Kajiya}{1986}]%
        {Kajiy1986}
\bibfield{author}{\bibinfo{person}{James~T. Kajiya}.}
  \bibinfo{year}{1986}\natexlab{}.
\newblock \showarticletitle{The Rendering Equation}.
\newblock \bibinfo{journal}{\emph{Computer Graphics (Proceedings of SIGGRAPH)}}
  \bibinfo{volume}{20}, \bibinfo{number}{4} (\bibinfo{date}{August}
  \bibinfo{year}{1986}), \bibinfo{pages}{143--150}.
\newblock
\urldef\tempurl%
\url{https://doi.org/10.1145/15886.15902}
\showDOI{\tempurl}


\bibitem[\protect\citeauthoryear{Kim, Park, Sohn, and Lin}{Kim
  et~al\mbox{.}}{2016}]%
        {KimPSL2016}
\bibfield{author}{\bibinfo{person}{Seungryong Kim}, \bibinfo{person}{Kihong
  Park}, \bibinfo{person}{Kwanghoon Sohn}, {and} \bibinfo{person}{Stephen
  Lin}.} \bibinfo{year}{2016}\natexlab{}.
\newblock \showarticletitle{Unified Depth Prediction and Intrinsic Image
  Decomposition from a Single Image via Joint Convolutional Neural Fields}. In
  \bibinfo{booktitle}{\emph{ECCV}}. \bibinfo{pages}{143--159}.
\newblock
\urldef\tempurl%
\url{https://doi.org/10.1007/978-3-319-46484-8_9}
\showDOI{\tempurl}


\bibitem[\protect\citeauthoryear{Kong, Gehler, and Black}{Kong
  et~al\mbox{.}}{2014}]%
        {KongGB2014}
\bibfield{author}{\bibinfo{person}{Naejin Kong}, \bibinfo{person}{Peter~V.
  Gehler}, {and} \bibinfo{person}{Michael~J. Black}.}
  \bibinfo{year}{2014}\natexlab{}.
\newblock \showarticletitle{Intrinsic Video}. In
  \bibinfo{booktitle}{\emph{ECCV}}. \bibinfo{pages}{360--375}.
\newblock
\urldef\tempurl%
\url{https://doi.org/10.1007/978-3-319-10605-2_24}
\showDOI{\tempurl}


\bibitem[\protect\citeauthoryear{Kovacs, Bell, Snavely, and Bala}{Kovacs
  et~al\mbox{.}}{2017}]%
        {KovacBSB2017}
\bibfield{author}{\bibinfo{person}{Balazs Kovacs}, \bibinfo{person}{Sean Bell},
  \bibinfo{person}{Noah Snavely}, {and} \bibinfo{person}{Kavita Bala}.}
  \bibinfo{year}{2017}\natexlab{}.
\newblock \showarticletitle{Shading Annotations in the Wild}. In
  \bibinfo{booktitle}{\emph{CVPR}}. \bibinfo{pages}{850--859}.
\newblock
\urldef\tempurl%
\url{https://doi.org/10.1109/CVPR.2017.97}
\showDOI{\tempurl}


\bibitem[\protect\citeauthoryear{Laffont, Bousseau, Paris, Durand, and
  Drettakis}{Laffont et~al\mbox{.}}{2012}]%
        {LaffoBPDD2012}
\bibfield{author}{\bibinfo{person}{Pierre-Yves Laffont},
  \bibinfo{person}{Adrien Bousseau}, \bibinfo{person}{Sylvain Paris},
  \bibinfo{person}{Frédo Durand}, {and} \bibinfo{person}{George Drettakis}.}
  \bibinfo{year}{2012}\natexlab{}.
\newblock \showarticletitle{Coherent Intrinsic Images from Photo Collections}.
\newblock \bibinfo{journal}{\emph{ACM Trans. Graph.}} \bibinfo{volume}{31},
  \bibinfo{number}{6} (\bibinfo{date}{November} \bibinfo{year}{2012}),
  \bibinfo{pages}{202:1--11}.
\newblock
\urldef\tempurl%
\url{https://doi.org/10.1145/2366145.2366221}
\showDOI{\tempurl}


\bibitem[\protect\citeauthoryear{Li, Dong, Peers, and Tong}{Li
  et~al\mbox{.}}{2017}]%
        {LiDPT2017}
\bibfield{author}{\bibinfo{person}{Xiao Li}, \bibinfo{person}{Yue Dong},
  \bibinfo{person}{Pieter Peers}, {and} \bibinfo{person}{Xin Tong}.}
  \bibinfo{year}{2017}\natexlab{}.
\newblock \showarticletitle{Modeling Surface Appearance from a Single
  Photograph Using Self-augmented Convolutional Neural Networks}.
\newblock \bibinfo{journal}{\emph{ACM Trans. Graph.}} \bibinfo{volume}{36},
  \bibinfo{number}{4} (\bibinfo{date}{July} \bibinfo{year}{2017}),
  \bibinfo{pages}{45:1--11}.
\newblock
\urldef\tempurl%
\url{https://doi.org/10.1145/3072959.3073641}
\showDOI{\tempurl}


\bibitem[\protect\citeauthoryear{Li, Xu, Ramamoorthi, Sunkavalli, and
  Chandraker}{Li et~al\mbox{.}}{2018}]%
        {LiXRSC2018}
\bibfield{author}{\bibinfo{person}{Zhengqin Li}, \bibinfo{person}{Zexiang Xu},
  \bibinfo{person}{Ravi Ramamoorthi}, \bibinfo{person}{Kalyan Sunkavalli},
  {and} \bibinfo{person}{Manmohan Chandraker}.}
  \bibinfo{year}{2018}\natexlab{}.
\newblock \showarticletitle{Learning to Reconstruct Shape and Spatially-varying
  Reflectance from a Single Image}.
\newblock \bibinfo{journal}{\emph{ACM Trans. Graph.}} \bibinfo{volume}{37},
  \bibinfo{number}{6} (\bibinfo{date}{November} \bibinfo{year}{2018}),
  \bibinfo{pages}{269:1--11}.
\newblock
\urldef\tempurl%
\url{https://doi.org/10.1145/3272127.3275055}
\showDOI{\tempurl}


\bibitem[\protect\citeauthoryear{Lin, Fisher, Dai, and Hanrahan}{Lin
  et~al\mbox{.}}{2017}]%
        {LinFDH2017}
\bibfield{author}{\bibinfo{person}{Sharon Lin}, \bibinfo{person}{Matthew
  Fisher}, \bibinfo{person}{Angela Dai}, {and} \bibinfo{person}{Pat Hanrahan}.}
  \bibinfo{year}{2017}\natexlab{}.
\newblock \bibinfo{title}{{LayerBuilder}: Layer Decomposition for Interactive
  Image and Video Color Editing}.  (\bibinfo{year}{2017}).
\newblock
\newblock
\shownote{\href{https://arxiv.org/abs/1701.03754}{arXiv:1701.03754}.}


\bibitem[\protect\citeauthoryear{Liu, Ceylan, Yumer, Yang, and Lien}{Liu
  et~al\mbox{.}}{2017}]%
        {LiuCYYL2017}
\bibfield{author}{\bibinfo{person}{Guilin Liu}, \bibinfo{person}{Duygu Ceylan},
  \bibinfo{person}{Ersin Yumer}, \bibinfo{person}{Jimei Yang}, {and}
  \bibinfo{person}{Jyh-Ming Lien}.} \bibinfo{year}{2017}\natexlab{}.
\newblock \showarticletitle{Material Editing Using a Physically Based Rendering
  Network}. In \bibinfo{booktitle}{\emph{ICCV}}. \bibinfo{pages}{2280--2288}.
\newblock
\urldef\tempurl%
\url{https://doi.org/10.1109/ICCV.2017.248}
\showDOI{\tempurl}


\bibitem[\protect\citeauthoryear{Lombardi and Nishino}{Lombardi and
  Nishino}{2016}]%
        {LombaN2016}
\bibfield{author}{\bibinfo{person}{Stephen Lombardi} {and} \bibinfo{person}{Ko
  Nishino}.} \bibinfo{year}{2016}\natexlab{}.
\newblock \showarticletitle{Reflectance and Illumination Recovery in the Wild}.
\newblock \bibinfo{journal}{\emph{IEEE Trans. Pattern Anal. Mach. Intell.}}
  \bibinfo{volume}{38}, \bibinfo{number}{1} (\bibinfo{date}{January}
  \bibinfo{year}{2016}), \bibinfo{pages}{129--141}.
\newblock
\urldef\tempurl%
\url{https://doi.org/10.1109/TPAMI.2015.2430318}
\showDOI{\tempurl}


\bibitem[\protect\citeauthoryear{Marschner and Greenberg}{Marschner and
  Greenberg}{1997}]%
        {MarscG1997}
\bibfield{author}{\bibinfo{person}{Stephen~R. Marschner} {and}
  \bibinfo{person}{Donald~P. Greenberg}.} \bibinfo{year}{1997}\natexlab{}.
\newblock \showarticletitle{Inverse lighting for photography}. In
  \bibinfo{booktitle}{\emph{Proceedings of the IS\&T Color Imaging
  Conference}}. \bibinfo{pages}{262--265}.
\newblock


\bibitem[\protect\citeauthoryear{Meka, Zollhöfer, Richardt, and Theobalt}{Meka
  et~al\mbox{.}}{2016}]%
        {MekaZRT2016}
\bibfield{author}{\bibinfo{person}{Abhimitra Meka}, \bibinfo{person}{Michael
  Zollhöfer}, \bibinfo{person}{Christian Richardt}, {and}
  \bibinfo{person}{Christian Theobalt}.} \bibinfo{year}{2016}\natexlab{}.
\newblock \showarticletitle{Live Intrinsic Video}.
\newblock \bibinfo{journal}{\emph{ACM Trans. Graph.}} \bibinfo{volume}{35},
  \bibinfo{number}{4} (\bibinfo{date}{July} \bibinfo{year}{2016}),
  \bibinfo{pages}{109:1--14}.
\newblock
\urldef\tempurl%
\url{https://doi.org/10.1145/2897824.2925907}
\showDOI{\tempurl}


\bibitem[\protect\citeauthoryear{Nam, Lee, Gutierrez, and Kim}{Nam
  et~al\mbox{.}}{2018}]%
        {NamLGK2018}
\bibfield{author}{\bibinfo{person}{Giljoo Nam}, \bibinfo{person}{Joo~Ho Lee},
  \bibinfo{person}{Diego Gutierrez}, {and} \bibinfo{person}{Min~H. Kim}.}
  \bibinfo{year}{2018}\natexlab{}.
\newblock \showarticletitle{Practical {SVBRDF} Acquisition of {3D} Objects with
  Unstructured Flash Photography}.
\newblock \bibinfo{journal}{\emph{ACM Trans. Graph.}} \bibinfo{volume}{37},
  \bibinfo{number}{6} (\bibinfo{date}{November} \bibinfo{year}{2018}),
  \bibinfo{pages}{267:1--12}.
\newblock
\urldef\tempurl%
\url{https://doi.org/10.1145/3272127.3275017}
\showDOI{\tempurl}


\bibitem[\protect\citeauthoryear{Narihira, Maire, and Yu}{Narihira
  et~al\mbox{.}}{2015}]%
        {NarihMY2015b}
\bibfield{author}{\bibinfo{person}{Takuya Narihira}, \bibinfo{person}{Michael
  Maire}, {and} \bibinfo{person}{Stella~X. Yu}.}
  \bibinfo{year}{2015}\natexlab{}.
\newblock \showarticletitle{Direct Intrinsics: Learning Albedo-Shading
  Decomposition by Convolutional Regression}. In
  \bibinfo{booktitle}{\emph{ICCV}}.
\newblock
\urldef\tempurl%
\url{https://doi.org/10.1109/ICCV.2015.342}
\showDOI{\tempurl}


\bibitem[\protect\citeauthoryear{Nayar, Krishnan, Grossberg, and Raskar}{Nayar
  et~al\mbox{.}}{2006}]%
        {NayarKGR2006}
\bibfield{author}{\bibinfo{person}{Shree~K. Nayar}, \bibinfo{person}{Gurunandan
  Krishnan}, \bibinfo{person}{Michael~D. Grossberg}, {and}
  \bibinfo{person}{Ramesh Raskar}.} \bibinfo{year}{2006}\natexlab{}.
\newblock \showarticletitle{Fast separation of direct and global components of
  a scene using high frequency illumination}.
\newblock \bibinfo{journal}{\emph{ACM Trans. Graph.}} \bibinfo{volume}{25},
  \bibinfo{number}{3} (\bibinfo{date}{July} \bibinfo{year}{2006}),
  \bibinfo{pages}{935--944}.
\newblock
\urldef\tempurl%
\url{https://doi.org/10.1145/1141911.1141977}
\showDOI{\tempurl}


\bibitem[\protect\citeauthoryear{Nestmeyer and Gehler}{Nestmeyer and
  Gehler}{2017}]%
        {NestmG2017}
\bibfield{author}{\bibinfo{person}{Thomas Nestmeyer} {and}
  \bibinfo{person}{Peter~V. Gehler}.} \bibinfo{year}{2017}\natexlab{}.
\newblock \showarticletitle{Reflectance Adaptive Filtering Improves Intrinsic
  Image Estimation}. In \bibinfo{booktitle}{\emph{CVPR}}.
  \bibinfo{pages}{1771--1780}.
\newblock
\urldef\tempurl%
\url{https://doi.org/10.1109/CVPR.2017.192}
\showDOI{\tempurl}


\bibitem[\protect\citeauthoryear{O'Toole, Mather, and Kutulakos}{O'Toole
  et~al\mbox{.}}{2016}]%
        {OToolMK2016}
\bibfield{author}{\bibinfo{person}{Matthew O'Toole}, \bibinfo{person}{John
  Mather}, {and} \bibinfo{person}{Kiriakos~N. Kutulakos}.}
  \bibinfo{year}{2016}\natexlab{}.
\newblock \showarticletitle{{3D} Shape and Indirect Appearance by Structured
  Light Transport}.
\newblock \bibinfo{journal}{\emph{IEEE Trans. Pattern Anal. Mach. Intell.}}
  \bibinfo{volume}{38}, \bibinfo{number}{7} (\bibinfo{date}{July}
  \bibinfo{year}{2016}), \bibinfo{pages}{1298--1312}.
\newblock
\urldef\tempurl%
\url{https://doi.org/10.1109/TPAMI.2016.2545662}
\showDOI{\tempurl}


\bibitem[\protect\citeauthoryear{Patow and Pueyo}{Patow and Pueyo}{2003}]%
        {PatowP2003}
\bibfield{author}{\bibinfo{person}{Gustavo Patow} {and} \bibinfo{person}{Xavier
  Pueyo}.} \bibinfo{year}{2003}\natexlab{}.
\newblock \showarticletitle{A Survey of Inverse Rendering Problems}.
\newblock \bibinfo{journal}{\emph{Comput. Graph. Forum}} \bibinfo{volume}{22},
  \bibinfo{number}{4} (\bibinfo{year}{2003}), \bibinfo{pages}{663--687}.
\newblock
\urldef\tempurl%
\url{https://doi.org/10.1111/j.1467-8659.2003.00716.x}
\showDOI{\tempurl}


\bibitem[\protect\citeauthoryear{Ramamoorthi and Hanrahan}{Ramamoorthi and
  Hanrahan}{2001}]%
        {RamamH2001}
\bibfield{author}{\bibinfo{person}{Ravi Ramamoorthi} {and} \bibinfo{person}{Pat
  Hanrahan}.} \bibinfo{year}{2001}\natexlab{}.
\newblock \showarticletitle{A signal-processing framework for inverse
  rendering}. In \bibinfo{booktitle}{\emph{SIGGRAPH}}.
  \bibinfo{pages}{117--128}.
\newblock
\urldef\tempurl%
\url{https://doi.org/10.1145/383259.383271}
\showDOI{\tempurl}


\bibitem[\protect\citeauthoryear{Ren, Dong, Lin, Tong, and Guo}{Ren
  et~al\mbox{.}}{2015}]%
        {RenDLTG2015}
\bibfield{author}{\bibinfo{person}{Peiran Ren}, \bibinfo{person}{Yue Dong},
  \bibinfo{person}{Stephen Lin}, \bibinfo{person}{Xin Tong}, {and}
  \bibinfo{person}{Baining Guo}.} \bibinfo{year}{2015}\natexlab{}.
\newblock \showarticletitle{Image Based Relighting Using Neural Networks}.
\newblock \bibinfo{journal}{\emph{ACM Trans. Graph.}} \bibinfo{volume}{34},
  \bibinfo{number}{4} (\bibinfo{date}{July} \bibinfo{year}{2015}),
  \bibinfo{pages}{111:1--12}.
\newblock
\urldef\tempurl%
\url{https://doi.org/10.1145/2766899}
\showDOI{\tempurl}


\bibitem[\protect\citeauthoryear{Richardt, Lopez-Moreno, Bousseau, Agrawala,
  and Drettakis}{Richardt et~al\mbox{.}}{2014}]%
        {RichaLBAD2014}
\bibfield{author}{\bibinfo{person}{Christian Richardt}, \bibinfo{person}{Jorge
  Lopez-Moreno}, \bibinfo{person}{Adrien Bousseau}, \bibinfo{person}{Maneesh
  Agrawala}, {and} \bibinfo{person}{George Drettakis}.}
  \bibinfo{year}{2014}\natexlab{}.
\newblock \showarticletitle{Vectorising Bitmaps into Semi-Transparent Gradient
  Layers}.
\newblock \bibinfo{journal}{\emph{Comput. Graph. Forum}} \bibinfo{volume}{33},
  \bibinfo{number}{4} (\bibinfo{date}{June} \bibinfo{year}{2014}),
  \bibinfo{pages}{11--19}.
\newblock
\urldef\tempurl%
\url{https://doi.org/10.1111/cgf.12408}
\showDOI{\tempurl}


\bibitem[\protect\citeauthoryear{Schultz and Hermes}{Schultz and
  Hermes}{2006}]%
        {SchulH2006}
\bibfield{author}{\bibinfo{person}{Christopher Schultz} {and}
  \bibinfo{person}{Thorsten Hermes}.} \bibinfo{year}{2006}\natexlab{}.
\newblock \bibinfo{booktitle}{\emph{Digital Keying Methods}}.
\newblock \bibinfo{type}{TZI-Bericht}~40.
  \bibinfo{institution}{Technologie-Zentrum Informatik, Bremen University}.
\newblock
\urldef\tempurl%
\url{http://www.tzi.de/fileadmin/resources/publikationen/tzi_berichte/TZI-Bericht-Nr._40.pdf}
\showURL{%
\tempurl}


\bibitem[\protect\citeauthoryear{Seitz, Matsushita, and Kutulakos}{Seitz
  et~al\mbox{.}}{2005}]%
        {SeitzMK2005}
\bibfield{author}{\bibinfo{person}{Steven~M. Seitz}, \bibinfo{person}{Yasuyuki
  Matsushita}, {and} \bibinfo{person}{Kiriakos~N. Kutulakos}.}
  \bibinfo{year}{2005}\natexlab{}.
\newblock \showarticletitle{A theory of inverse light transport}. In
  \bibinfo{booktitle}{\emph{ICCV}}, Vol.~\bibinfo{volume}{2}.
  \bibinfo{pages}{1440--1447}.
\newblock
\urldef\tempurl%
\url{https://doi.org/10.1109/ICCV.2005.25}
\showDOI{\tempurl}


\bibitem[\protect\citeauthoryear{Shen, Yan, Chen, Sun, and Li}{Shen
  et~al\mbox{.}}{2014}]%
        {ShenYCSL2014}
\bibfield{author}{\bibinfo{person}{Jianbing Shen}, \bibinfo{person}{Xing Yan},
  \bibinfo{person}{Lin Chen}, \bibinfo{person}{Hanqiu Sun}, {and}
  \bibinfo{person}{Xuelong Li}.} \bibinfo{year}{2014}\natexlab{}.
\newblock \showarticletitle{Re-texturing by intrinsic video}.
\newblock \bibinfo{journal}{\emph{Information Sciences}}  \bibinfo{volume}{281}
  (\bibinfo{date}{October} \bibinfo{year}{2014}), \bibinfo{pages}{726--735}.
\newblock
\urldef\tempurl%
\url{https://doi.org/10.1016/j.ins.2014.02.134}
\showDOI{\tempurl}


\bibitem[\protect\citeauthoryear{Shen, Yang, Jia, and Li}{Shen
  et~al\mbox{.}}{2011}]%
        {ShenYJL2011}
\bibfield{author}{\bibinfo{person}{Jianbing Shen}, \bibinfo{person}{Xiaoshan
  Yang}, \bibinfo{person}{Yunde Jia}, {and} \bibinfo{person}{Xuelong Li}.}
  \bibinfo{year}{2011}\natexlab{}.
\newblock \showarticletitle{Intrinsic images using optimization}. In
  \bibinfo{booktitle}{\emph{CVPR}}. \bibinfo{pages}{3481--3487}.
\newblock
\urldef\tempurl%
\url{https://doi.org/10.1109/CVPR.2011.5995507}
\showDOI{\tempurl}


\bibitem[\protect\citeauthoryear{Shi, Dong, Su, and Yu}{Shi
  et~al\mbox{.}}{2017}]%
        {ShiDSY2017}
\bibfield{author}{\bibinfo{person}{Jian Shi}, \bibinfo{person}{Yue Dong},
  \bibinfo{person}{Hao Su}, {and} \bibinfo{person}{Stella~X. Yu}.}
  \bibinfo{year}{2017}\natexlab{}.
\newblock \showarticletitle{Learning Non-{Lambertian} Object Intrinsics across
  {ShapeNet} Categories}. In \bibinfo{booktitle}{\emph{CVPR}}.
  \bibinfo{pages}{5844--5853}.
\newblock
\urldef\tempurl%
\url{https://doi.org/10.1109/CVPR.2017.619}
\showDOI{\tempurl}


\bibitem[\protect\citeauthoryear{Tan, Echevarria, and Gingold}{Tan
  et~al\mbox{.}}{2018}]%
        {TanEG2018a}
\bibfield{author}{\bibinfo{person}{Jianchao Tan}, \bibinfo{person}{Jose
  Echevarria}, {and} \bibinfo{person}{Yotam Gingold}.}
  \bibinfo{year}{2018}\natexlab{}.
\newblock \showarticletitle{Efficient Palette-based Decomposition and
  Recoloring of Images via {RGBXY}-space Geometry}.
\newblock \bibinfo{journal}{\emph{ACM Trans. Graph.}} \bibinfo{volume}{37},
  \bibinfo{number}{6} (\bibinfo{date}{November} \bibinfo{year}{2018}),
  \bibinfo{pages}{262:1--10}.
\newblock
\urldef\tempurl%
\url{https://doi.org/10.1145/3272127.3275054}
\showDOI{\tempurl}


\bibitem[\protect\citeauthoryear{Tan, Lien, and Gingold}{Tan
  et~al\mbox{.}}{2016}]%
        {TanLG2016}
\bibfield{author}{\bibinfo{person}{Jianchao Tan}, \bibinfo{person}{Jyh-Ming
  Lien}, {and} \bibinfo{person}{Yotam Gingold}.}
  \bibinfo{year}{2016}\natexlab{}.
\newblock \showarticletitle{Decomposing Images into Layers via {RGB}-space
  Geometry}.
\newblock \bibinfo{journal}{\emph{ACM Trans. Graph.}} \bibinfo{volume}{36},
  \bibinfo{number}{1} (\bibinfo{date}{November} \bibinfo{year}{2016}),
  \bibinfo{pages}{7:1--14}.
\newblock
\urldef\tempurl%
\url{https://doi.org/10.1145/2988229}
\showDOI{\tempurl}


\bibitem[\protect\citeauthoryear{Wu, Zollhöfer, Nießner, Stamminger, Izadi,
  and Theobalt}{Wu et~al\mbox{.}}{2014}]%
        {WuZNSIT2014}
\bibfield{author}{\bibinfo{person}{Chenglei Wu}, \bibinfo{person}{Michael
  Zollhöfer}, \bibinfo{person}{Matthias Nießner}, \bibinfo{person}{Marc
  Stamminger}, \bibinfo{person}{Shahram Izadi}, {and}
  \bibinfo{person}{Christian Theobalt}.} \bibinfo{year}{2014}\natexlab{}.
\newblock \showarticletitle{Real-time Shading-based Refinement for Consumer
  Depth Cameras}.
\newblock \bibinfo{journal}{\emph{ACM Trans. Graph.}} \bibinfo{volume}{33},
  \bibinfo{number}{6} (\bibinfo{date}{November} \bibinfo{year}{2014}),
  \bibinfo{pages}{200:1--10}.
\newblock
\urldef\tempurl%
\url{https://doi.org/10.1145/2661229.2661232}
\showDOI{\tempurl}


\bibitem[\protect\citeauthoryear{Wu, Wang, and Zhou}{Wu et~al\mbox{.}}{2016}]%
        {WuWZ2016}
\bibfield{author}{\bibinfo{person}{Hongzhi Wu}, \bibinfo{person}{Zhaotian
  Wang}, {and} \bibinfo{person}{Kun Zhou}.} \bibinfo{year}{2016}\natexlab{}.
\newblock \showarticletitle{Simultaneous Localization and Appearance Estimation
  with a Consumer {RGB-D} Camera}.
\newblock \bibinfo{journal}{\emph{IEEE Trans. Vis. Comput. Graph.}}
  \bibinfo{volume}{22}, \bibinfo{number}{8} (\bibinfo{date}{August}
  \bibinfo{year}{2016}), \bibinfo{pages}{2012--2023}.
\newblock
\urldef\tempurl%
\url{https://doi.org/10.1109/TVCG.2015.2498617}
\showDOI{\tempurl}


\bibitem[\protect\citeauthoryear{Ye, Garces, Liu, Dai, and Gutierrez}{Ye
  et~al\mbox{.}}{2014}]%
        {YeGLDG2014}
\bibfield{author}{\bibinfo{person}{Genzhi Ye}, \bibinfo{person}{Elena Garces},
  \bibinfo{person}{Yebin Liu}, \bibinfo{person}{Qionghai Dai}, {and}
  \bibinfo{person}{Diego Gutierrez}.} \bibinfo{year}{2014}\natexlab{}.
\newblock \showarticletitle{Intrinsic Video and Applications}.
\newblock \bibinfo{journal}{\emph{ACM Trans. Graph.}} \bibinfo{volume}{33},
  \bibinfo{number}{4} (\bibinfo{date}{July} \bibinfo{year}{2014}),
  \bibinfo{pages}{80:1--11}.
\newblock
\urldef\tempurl%
\url{https://doi.org/10.1145/2601097.2601135}
\showDOI{\tempurl}


\bibitem[\protect\citeauthoryear{Yu, Debevec, Malik, and Hawkins}{Yu
  et~al\mbox{.}}{1999}]%
        {YuDMH1999}
\bibfield{author}{\bibinfo{person}{Yizhou Yu}, \bibinfo{person}{Paul Debevec},
  \bibinfo{person}{Jitendra Malik}, {and} \bibinfo{person}{Tim Hawkins}.}
  \bibinfo{year}{1999}\natexlab{}.
\newblock \showarticletitle{Inverse global illumination: recovering reflectance
  models of real scenes from photographs}. In
  \bibinfo{booktitle}{\emph{SIGGRAPH}}. \bibinfo{pages}{215--224}.
\newblock
\urldef\tempurl%
\url{https://doi.org/10.1145/311535.311559}
\showDOI{\tempurl}


\bibitem[\protect\citeauthoryear{Zhou, Krähenbühl, and Efros}{Zhou
  et~al\mbox{.}}{2015}]%
        {ZhouKE2015}
\bibfield{author}{\bibinfo{person}{Tinghui Zhou}, \bibinfo{person}{Philipp
  Krähenbühl}, {and} \bibinfo{person}{Alyosha Efros}.}
  \bibinfo{year}{2015}\natexlab{}.
\newblock \showarticletitle{Learning Data-driven Reflectance Priors for
  Intrinsic Image Decomposition}. In \bibinfo{booktitle}{\emph{ICCV}}.
  \bibinfo{pages}{3469--3477}.
\newblock
\urldef\tempurl%
\url{https://doi.org/10.1109/ICCV.2015.396}
\showDOI{\tempurl}


\bibitem[\protect\citeauthoryear{Zollhöfer, Nießner, Izadi, Rhemann, Zach,
  Fisher, Wu, Fitzgibbon, Loop, Theobalt, and Stamminger}{Zollhöfer
  et~al\mbox{.}}{2014}]%
        {ZollhNIRZFWFLTS2014}
\bibfield{author}{\bibinfo{person}{Michael Zollhöfer},
  \bibinfo{person}{Matthias Nießner}, \bibinfo{person}{Shahram Izadi},
  \bibinfo{person}{Christoph Rhemann}, \bibinfo{person}{Christopher Zach},
  \bibinfo{person}{Matthew Fisher}, \bibinfo{person}{Chenglei Wu},
  \bibinfo{person}{Andrew Fitzgibbon}, \bibinfo{person}{Charles Loop},
  \bibinfo{person}{Christian Theobalt}, {and} \bibinfo{person}{Marc
  Stamminger}.} \bibinfo{year}{2014}\natexlab{}.
\newblock \showarticletitle{Real-time Non-rigid Reconstruction Using an {RGB-D}
  Camera}.
\newblock \bibinfo{journal}{\emph{ACM Trans. Graph.}} \bibinfo{volume}{33},
  \bibinfo{number}{4} (\bibinfo{date}{July} \bibinfo{year}{2014}),
  \bibinfo{pages}{156:1--12}.
\newblock
\urldef\tempurl%
\url{https://doi.org/10.1145/2601097.2601165}
\showDOI{\tempurl}


\bibitem[\protect\citeauthoryear{Zoran, Isola, Krishnan, and Freeman}{Zoran
  et~al\mbox{.}}{2015}]%
        {ZoranIKF2015}
\bibfield{author}{\bibinfo{person}{Daniel Zoran}, \bibinfo{person}{Phillip
  Isola}, \bibinfo{person}{Dilip Krishnan}, {and} \bibinfo{person}{William~T.
  Freeman}.} \bibinfo{year}{2015}\natexlab{}.
\newblock \showarticletitle{Learning Ordinal Relationships for Mid-Level
  Vision}. In \bibinfo{booktitle}{\emph{ICCV}}. \bibinfo{pages}{388--396}.
\newblock
\urldef\tempurl%
\url{https://doi.org/10.1109/ICCV.2015.52}
\showDOI{\tempurl}


\end{thebibliography}

\end{document}